\documentclass{article}

\usepackage{amsmath}
\usepackage{amsfonts}
\usepackage{multirow}
\usepackage{graphicx}
\usepackage{caption}
\usepackage{subcaption}
\usepackage{longtable}
\usepackage{pdflscape}
\usepackage[a4paper, margin=0.7in]{geometry}
\usepackage{bm}
\usepackage{afterpage}
\usepackage{multicol}
\usepackage{float}
\usepackage{longtable}
\usepackage[toc,page]{appendix}
\usepackage{breqn}
\usepackage{lscape}
\usepackage{amssymb,mathtools}
\usepackage{rotating}
\usepackage{authblk}\usepackage{siunitx}
\usepackage[backend=bibtex, minbibnames=3,style=numeric,citestyle=numeric-comp,sorting=none, giveninits ]{biblatex}
\DeclareNameAlias{author}{last-first}

\usepackage{xcolor}

\DeclareMathOperator*{\argmin}{arg\,min}

\usepackage{titlesec}
\usepackage{enumitem}
\usepackage{hyperref}

\titleformat{\paragraph}[runin]
{\normalfont\normalsize\bfseries}{\theparagraph}{1em}{}[:]
\titlespacing*{\paragraph}
{0pt}{3.25ex plus 1ex minus .2ex}{1.5ex plus .2ex}

\providecommand{\keywords}[1]{\textbf{\textit{Keywords---}} #1}

\DeclareMathAlphabet\mathbfcal{OMS}{cmsy}{b}{n}

\title{A proof of concept study for machine learning application to stenosis detection}
\author[1]{Gareth Jones}
\author[2]{Jim Parr}
\author[1]{Perumal Nithiarasu}
\author[1, $\dagger$]{Sanjay Pant}
\affil[1]{College of Engineering, Swansea University, Swansea, United Kingdom.}
\affil[2]{McLaren Technology Centre, Woking, United Kingdom.}

\affil[$\dagger$]{Corresponding author: Sanjay.Pant@swansea.ac.uk}

\date{}      
\bibliography{article.bib}

\begin{document}
\maketitle

\begin{abstract}
\noindent This proof of concept (PoC)  assesses the ability of machine learning (ML) classifiers to predict the presence of a stenosis in a  three vessel arterial system consisting of the abdominal aorta bifurcating into the two common iliacs. A virtual patient database (VPD) is created using one-dimensional pulse wave propagation model of haemodynamics. Four different machine learning (ML) methods are used to train and test a series of classifiers---both binary and multiclass---to distinguish between healthy and unhealthy virtual patients (VPs) using different combinations of pressure and flow-rate measurements. 
It is found that the ML classifiers achieve specificities larger than 80\% and sensitivities ranging from 50--75\%. The most balanced classifier also achieves an area under the receiver operative characteristic curve of 0.75, outperforming approximately 20 methods used in clinical practice, and thus placing the method as moderately accurate.
Other important observations from this study are that: i) few measurements can provide similar classification accuracies compared to the case when more/all the measurements are used; ii) some measurements are more informative than others for classification; and iii) a modification of standard methods can result in detection of not only the presence of stenosis, but also the stenosed vessel. 
\\[3pt]

\noindent \keywords{Arterial disease diagnosis, machine learning, virtual patient database, pulse wave haemodynamics}

\end{abstract}

\section{Introduction}

While there are many forms of arterial disease, one of the most common is stenosis, which refers to the 
narrowing of an arterial vessel. This  is normally caused by a build up of fatty deposits, known as atherosclerosis. Stenosis can be be categorised into several sub-diseases depending on its location. Three of the most common forms of stenosis are peripheral artery disease (PAD), carotid artery stenosis, and subclavian artery stenosis (SS). The prevalence of PAD and SS have been recorded to vary between 1.9\% and 18.83\% within different demographics \cite{fowkes2013comparison, shadman2004subclavian}, while carotid artery stenosis has been recorded to affect 3.8\% of men and 2.7\% of women \cite{mathiesen2001prevalence}.

Current methods for the detection of arterial disease are primarily based on imaging techniques \cite{titi2007comparison, litt1991diagnosis, quill1989ultrasonic, leopold1972ultrasonic}, and so are often impractical for large scale screening, expensive, or both. If a new inexpensive and non-invasive method for the detection of stenosis is found, the cost effectiveness of large scale screening could be improved making both continuous monitoring and screening feasible. One such alternative is to use easily acquirable  pressure and flow-rate measurements at accessible peripheral locations within the circulatory system and use them for diagnosis. It is known from the principles of fluid mechanics that if the cross sectional area of a vessel is changed, the pressure and flow-rate profiles of fluid passing through that vessel will also change \cite{may1963hemodynamic, donohue1993assessing, ujiie1999effects, sazonov2017novel}. Applying this to arterial disease, the inclusion of a stenosis within a patients arterial network may create detectable biomarkers within the pressure and flow-rate profiles of blood. This precise hypothesis is explored in this study.

A previous study \cite{sazonov2017novel} has  explored the use of physics based models of pulsewave propagation to predict the presence of an aneurysm, another common form of arterial disease, using flow-rate measurements. Its use for disease detection is, however, limited by the the need for patient specific parameters. If a consistent and significant biomarker of arterial disease is found within pressure and flow-rate profiles, irrespective of a patients individual arterial network, it would be possible to predict the presence of a stenosis using only these measurements. This would allow for inexpensive and non-invasive screening of patients for arterial disease. As opposed to a mechanistic approach to such an inverse problem, this study explores a pure data-driven machine learning approach for finding such biomarkers.

It is likely that the indicative biomarkers of arterial disease held within pressure and flow-rate profiles consist of micro inter- and intra-measurement details. Discovery of these biomarkers through a traditional hypothesis driven scientific method \cite{voit2019perspective} and a classical inverse problems approach is difficult. If a large database of pressure and flow-rate measurements taken from patients of known arterial health is available, it maybe possible for a machine learning (ML) classifier to be trained to not only discover but also exploit any biomarkers within the pressure and flow-rate profiles.

The aim of this proof of concept (PoC) study is to carry out an initial investigation into the potential of using ML classification algorithms to predict the presence of stenosis, using haemodynamics measurements. To train and test such ML classifiers, a large database of measurements taken from patients of known arterial health is required. As opposed to using measurements  from a real population, which are unavailable, a synthetic virtual patient database (VPD), similar to that presented in \cite{willemet2015database}, is created through the use of a physics based model of pulse wave propagation. To create the VPD, \textit{a priori} distributions are first constructed for the parameters describing the arterial networks of virtual patients (VPs) across the resulting VPD. Random realisations are then sampled  from these distributions, and the physics based model is solved to obtain the corresponding pressure and flow-rate profiles. 
Finally, ``Hard'' filters, \textit{i.e.} the direct imposition of bounds on the ranges of pressure profiles, are applied to the VPD to reduce the occurrence of physiologically unrealistic VPs.

This virtual population is then used to train and test a series of ML classifiers to detect arterial disease, and test their performance. Focus is on assessing feasibility and uncovering behaviours and patterns in the performance of classification methods, rather than optimisation and creation of increasingly complex ML models for maximum accuracy. 
Understanding the behaviour of classifiers will allow subsequent, more complex, studies to leverage on these observations.

In what follows, first the design of the VPD---its motivation,  physics based model,  the arterial network, its parameterisation, probability distributions, and filters---is presented. This is followed by the ML setup, its relation to the size of the VPD required, brief description of the ML methods and metrics to quantify their performance. Finally, the results and analysis of the ML methods performance are presented, with a focus on uncovering why some ML methods perform better than others and which measurements (and their combinations) are more informative.

\section{Virtual patient database}
\label{section_VPD_creation}

\subsection{Motivation}
\label{section_motivation_VPD}

To train and test ML classifiers a large database of haemodynamics measurements taken from a comprehensive cohort of patients is required. The corresponding correct arterial health of these patients is also required. As opposed to using measurements taken from real patients, VPs are created using a physics based model of pulse wave propagation. This VP approach has several advantages:
\begin{enumerate}[leftmargin=*]
\item \textbf{Expense}: creating VPs is relatively inexpensive. The primary cost associated with the creation of VPs is computational, and thus negligible in comparison to data acquisition in a real population.
\item \textbf{Class imbalance}: creating VPs allows for the control of the distribution of different diseased states. For example, in a real population the rate of arterial disease can vary between 1\% and 20\%. During the creation of VPs, however, 50\% diseased patients can be created to ensure a balanced data set.
\item \textbf{Measurement availability}: using VPs allows for measurements of pressure and flow-rate to be taken at any location within the arterial system. This allows for an \emph{a priori} assessment of ML classifiers using all possible combinations of pressure and flow-rate measurements.
\end{enumerate}

\noindent While there are limitations to the measurements that can be non-invasively and inexpensively obtained for a clinical application, pressure and flow-rate measurements throughout the arterial network are useful as they allow the impact of measurement location on performance to be investigated. This benefit is particularly important for this PoC where feasibility of the ML approach is being assessed. A  primary purpose of this study is to gain an understanding of the patterns between the measurements and classification accuracy.

\subsection{Physics based model of pulse wave propagation}
\label{section_1D_model}

To compute the pressure and flow-rate waveforms associated with VPs, a physics based model of one-dimensional pulse wave propagation is adopted \cite{boileau2015benchmark}. 
By considering each vessel within the network to be a deforming tube, a system of two governing equations can be derived. These equations represent  conservation of mass and momentum balance with the assumption that blood is incompressible and that the tube walls are impermeable. The system of equations is (see \cite{alastruey2012arterial} for details):
\begin{equation}
\frac{\partial{A}}{\partial{t}}+\frac{\partial{(UA)}}{\partial{x}}=0,
\label{eq_mass_conservation}
\end{equation}
\begin{equation}
\frac{\partial{U}}{\partial{t}}+U\frac{\partial{U}}{\partial{x}}+\frac{1}{\rho}\frac{\partial{P}}{\partial{x}}=\frac{f}{\rho{A}},
\label{eq_momentum_conservation}
\end{equation}
where $P(x, t)$, $U(x, t)$, and  $A(x, t)$ represent the pressure, flow velocity, and arterial cross-sectional area, respectively, at spatial coordinate $x$ and time $t$; $\rho$ and $\mu$ represent the density and the dynamic viscosity of blood, respectively; and   $f$ represents the frictional force per unit length described as follows
\begin{equation}
f(x, t)=-2({\zeta}+2){\mu}{\pi}U,
\label{eq_f}
\end{equation}
where $\zeta$ is a constant that depends on the velocity profile across the arterial cross-section. To close this system of equations, a mechanical model of the displacement of the vessel walls \cite{boileau2015benchmark} is included:
%
%
%
\begin{equation}
P-P_{ext}=P_d+\beta\frac{\sqrt{A}-\sqrt{A_d}}{A_d},
\label{eq_pressure_area}
\end{equation}
with 
\begin{equation}
\beta=\frac{4}{3}Eh\sqrt{\pi},
\label{eq_beta}
\end{equation}
where $P_{ext}$ represents the external pressure, $P_{d}$ represents the diastolic blood pressure, $A_d$ represents the diastolic area of the vessel, $E$ represents the vessel wall's Young's modulus, and $h$ represents the vessel wall's thickness. This system of equations has been previously used and tested extensively \cite{boileau2015benchmark, formaggia2003one, alastruey2012arterial, olufsen2000numerical, reymond2009validation, matthys2007pulse}.\\

\begin{figure}[tb]
\centering
\includegraphics[width=5.5in]{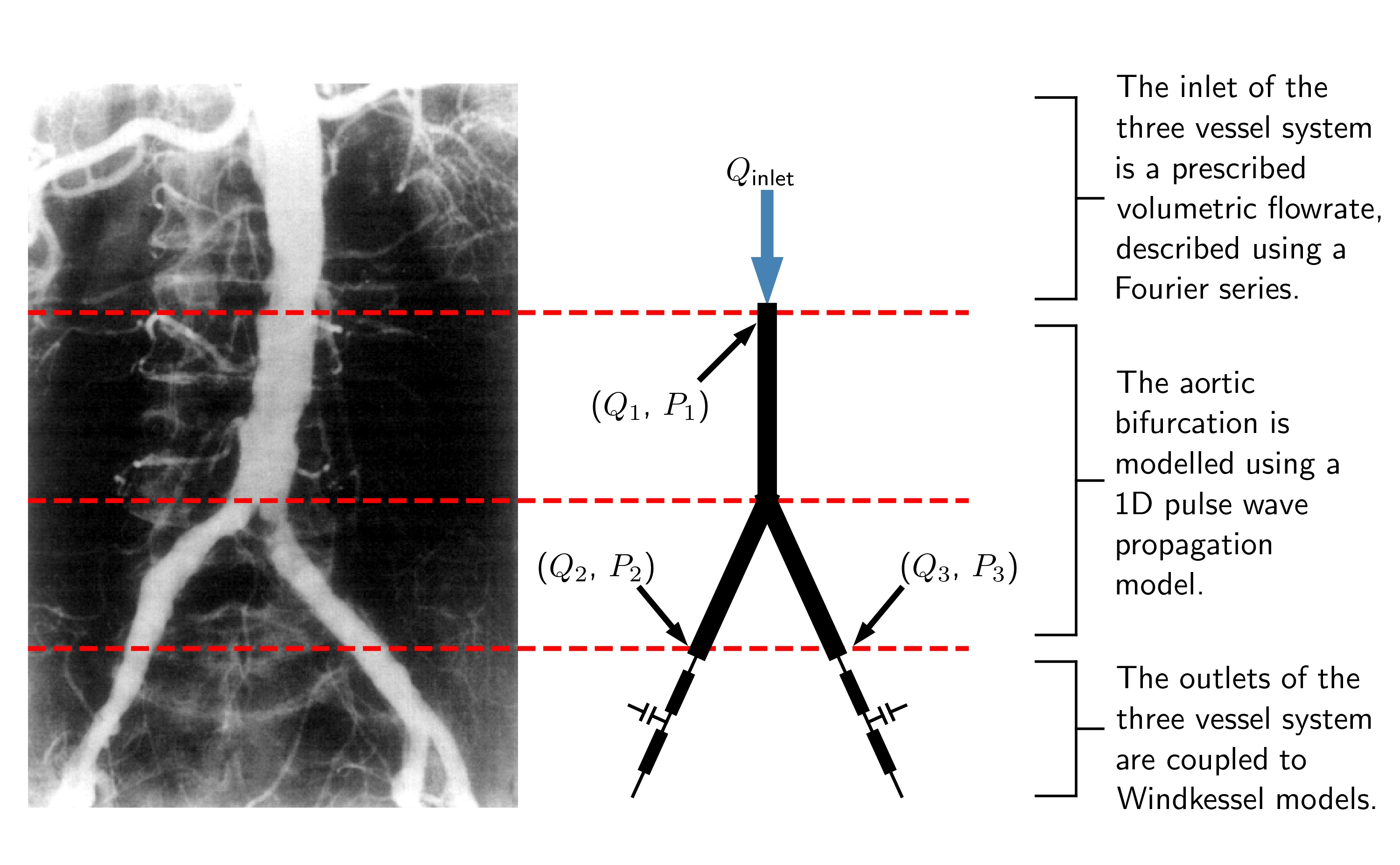}
\caption{The inlet and outlet boundary conditions to the  model. The relation of the model to the aortic bifurcation is also shown through comparison to an angiogram (reprinted from \cite{burnham1992noninvasive} with permission from Elsevier)}
\label{figure_model_overview}
\end{figure}

\subsection{Arterial network}
\label{section_arterial_network}

In this study, the network of interest is the abdominal aorta bifurcating into the two common iliacs. A pre-existing model for this is taken as a reference network from \cite{boileau2015benchmark}. This is shown in Figure \ref{figure_model_overview}, where the three vessels (abdominal aorta and the two iliacs) are represented in 1D while suitable boundary conditions are imposed at the inlet and the outlets.

\begin{figure}[tb]
\centering 
\includegraphics[width=2.5in]{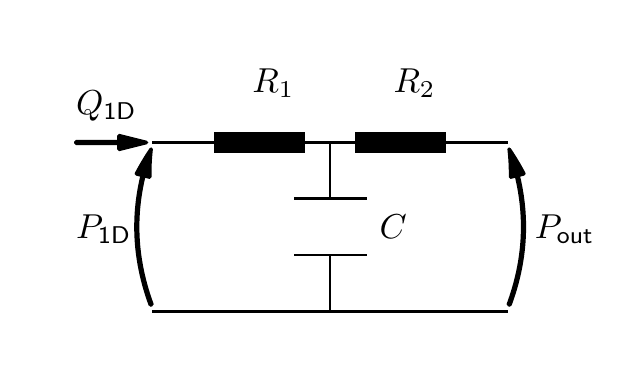}
\caption{The configuration of a three element Windkessel model: $Q_{\text{1D}}$ and $P_{\text{1D}}$ represent the volumetric flow-rate and pressure, respectively, at the terminal boundary of the 1D system.}
\label{figure_windkessel}
\end{figure}

At the inlet a time varying volumetric flow-rate is prescribed. The terminal outlets are coupled to three element Windkessel models \cite{westerhof2009arterial}, which replicate the effect of peripheral arteries. Each Windkessel model, as shown in Figure \ref{figure_windkessel}, consists of two resistors, $R_1$ and $R_2$, which represent the viscous resistances of the large arteries and the micro-vascular system, respectively, and a capacitor $C$, which  represents the compliance of large arteries.

\subsection{Numerical scheme}
\label{section_numerical_scheme}

With the specification of the network mechanical parameters and the boundary conditions, the model of section \ref{section_1D_model} is solved to compute the pressure and flow-rate waveforms across the network.
The system of equations is numerically solved using a discontinuous Galerkin scheme, see \cite{alastruey2012arterial} for details. This scheme is chosen as a pre-existing solver is available that has been successfully validated against a 3D model of blood flow through stenosed arterial vessels \cite{boileau2018estimating}.

\subsection{Parameterisation of the arterial network}
\label{section_topology}

This section presents the parameterisation of the arterial network for the creation of VPs. Once parameterised, the network parameters can be randomly sampled to create VPs.
The inlet volumetric flow-rate profile, $Q_{\text{inlet}}(t)$,  is described using a Fourier series (FS) representation:
\begin{equation}
Q_{\text{inlet}}(t)=\sum_{n=0}^N a_n \sin (n \omega t) + b_n \cos(n \omega t),
\label{eq_FS_rep_1}
\end{equation}
where $a_n$ and $b_n$ represent the $n^{\text{th}}$ sine and cosine FS coefficients, respectively; $N$ represents the truncation order; and $\omega={2 \pi}/{T}$, with $T$ as the time period of the cardiac cycle. It is found that the time domain inlet flow-rate profile of \cite{boileau2015benchmark} can be described to a high level of precision using a FS truncated at the $5^{\text{th}}$ order. Thus, the time domain inlet flow-rate profile is described by:
\begin{equation}
\bm{Q}_{\text{inlet}}=\{a_0=0, b_0, a_1, b_1,...,a_5, b_5\},
\label{eq_FS_vector_1}
\end{equation} 
requiring specification of 11 coefficients.

Since the three vessel segments in the network are short, It is assumed that the properties of all the three vessels are constant along their lengths. To impose geometric and mechanical symmetry on the lower extremities, the two common iliac arteries are assumed to share identical properties. This symmetry, however, is not extended to the terminal Windkessel model parameters.  The parameterisation of the network thus requires specification of the following 25 parameters:
\begin{itemize}[leftmargin=*]
\item \textbf{Six geometric properties}: the two common iliac arteries require specification of a single length, a reference area, and a wall thickness. These three properties are also required for the aorta.
\item \textbf{Two mechanical properties}: the Young's modulus of the aorta and the common iliacs needs to be specified.
\item \textbf{Six terminal boundary parameters}: each of the Windkessel models require two resistances and a compliance.
\item \textbf{11 FS coefficients}: the time domain inlet flow-rate profile is described using a FS truncated at the $5^{\text{th}}$ order.
\end{itemize}    

\begin{figure}[tb]
\centering
\includegraphics[width=2.5in]{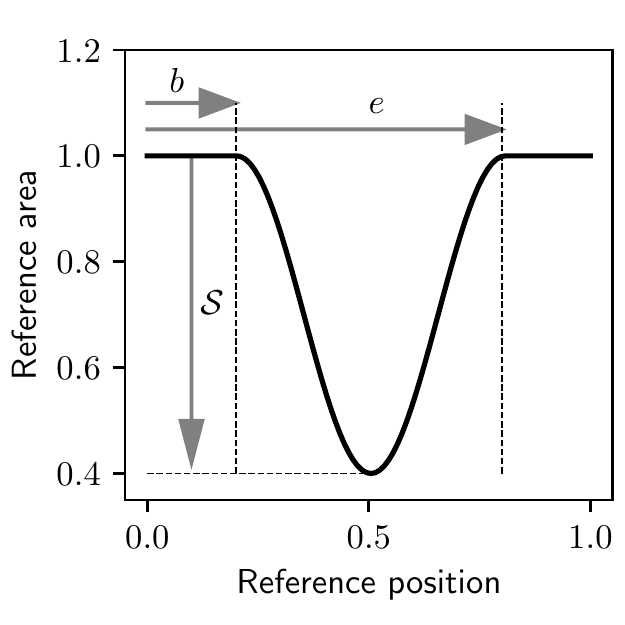}
\caption{An example of a 60\% stenosis with a start location of $b=0.2$ and an end location of $e=0.8$.}
\label{figure_area_reduction}
\end{figure}

For an ML classifier to be trained to distinguish between healthy and unhealthy patients, examples of both classifications are required within the VPD. A parameterisation must, therefore, be chosen to describe stenosed arterial vessels. For simplicity, all VPs are limited to having a maximum of one diseased vessel. To create a change in the reference area of a diseased vessel a normalised map of each vessel's area is produced. Both the length and cross sectional area of the vessel is normalised between 0 and 1. This map, for a 60\% stenosis, is shown in  Figure \ref{figure_area_reduction}, where the x-axis  represents the reference position along the length of the vessel and the y-axis represents the reference cross sectional area. For healthy vessels the normalised reference cross sectional area is constant and equal to 1. For unhealthy vessels a cosine curve is used to create a change in area. This cosine curve is scaled using three parameters to create variation in location and severity of disease between patients. These parameters are the severity $\mathcal{S}$, the start location $b$, and the end location $e$ of the disease. The normalised cross sectional area $A_n$ of a diseased vessel at a spatial location  $x$ is described as: 
\begin{equation}
A_{n}\!=\! 
\begin{cases}
\bigg(1\! - \! \dfrac{\mathcal{S}}{2} \bigg) +  \dfrac{S}{2} \cos \left(\dfrac{2 (x_n-b) \pi}{e-b}\right) & \text{for } b\leq x_n \leq e \\
\phantom{x} 1 & \text{otherwise.}
\end{cases}
\label{eq_area_profile}
\end{equation}
%
Thus, in addition to the 25 parameters for the description of a healthy subject, three more parameters are required for specification of disease. Random realisations of these parameters are sampled and the physics based model of pulse wave propagation is solved to produce each VP. The probability distributions of these parameters are described next.

\subsection{Probability distributions}
Ideally the distribution of both arterial network parameters and the resulting pressure and flow-rate profiles should be representative of those measured in a real population. Since one dimensional arterial network parameters are generally either expensive and invasive to obtain or non-physical (so cannot be directly measured), their exact distributions are not known. Thus, \textit{A priori} distributions are assumed for both healthy and diseased virtual subjects, as described next.
\subsubsection{Healthy subjects:} \textit{A priori} distributions are  assumed for the arterial network parameters, based on values reported in literature \cite{boileau2015benchmark}. It is assumed that across a large population all parameters required to describe VPs arterial networks, excluding the disease parameters, are independent and normally distributed. The mean value for each of these parameters is taken from \cite{boileau2015benchmark} and the standard deviation is set to be 20\% of the mean, as summarised in Table \ref{table_parameter_values}. VPs are assigned disease so that the VPD consists of an expected 50\% healthy patients, and there is an expected equal number of aortic, first iliac, and second iliac stenosis VPs.\\

 \subsubsection{Diseased patients:}
 In addition to the parameters described for healthy patients above, a diseased patient is characterised by three more parameters---
 disease severity, start location, and end location---which are assigned uniform distributions based on physical constraints. A fourth parameter, the reference location of the disease (represented by $r$) is introduced. This parameter is included to impose a minimum stenosis length of 10\% of the vessel length. The four parameters are sequentially sampled from uniform distributions within the following bounds:
\begin{equation}
\mathrm{Constraints: }\;
\begin{cases}
0.2 \leq r \leq 0.8\\
0.1 \leq b \leq r-0.05 \\
r+0.05 \leq e \leq 0.9 \\
0.5 \leq \mathcal{S} \leq 0.9.\\
\end{cases}
\end{equation}

\begin{table*}
\begin{center}
\def\arraystretch{1.2}
\begin{tabular}{|c|c c |c c |}
\hline
\textbf{Parameter} & \multicolumn{2}{c|}{\textbf{Mean}} & \multicolumn{2}{c|}{\textbf{Standard deviation}} \\
&  \textbf{Aorta} & \textbf{Iliac} & \textbf{Aorta} & \textbf{Iliac} \\
\hline
Length & 8.6cm & 8.5cm & 1.72cm & 1.7cm \\
Wall thickness & 1.03mm & 0.72mm & 0.21mm & 0.14mm \\
Reference diameter &  1.72cm & 1.2cm & 0.344cm & 0.24cm \\
Young's modulus & 500kPa & 700kPa & 100kPa & 140kPa \\
$R_1$ & - & $6.81{\times}10^7$ $Pa$ $s$ $m^{-3}$ & - & $1.36{\times}10^6$ $Pa$ $s$ $m^{-3}$ \\
$R_2$ & - & $3.10{\times}10^9$ $Pa$ $s$ $m^{-3}$ & - & $6.20{\times}10^8$ $Pa$ $s$ $m^{-3}$ \\
$C$ & - & $3.67{\times}10^{-10}$ $m^3$ $Pa^{-1}$ & - & $7.33{\times}10^{-11}$ $m^3$ $Pa^{-1}$ \\
\hline
\end{tabular}
\caption{Mean and standard deviations of the arterial network parameters.}
\label{table_parameter_values}
\end{center}
\end{table*}
The assumption that all arterial network parameters are independent and normally distributed is likely physiologically incorrect. To correct for this assumption, post simulation filters are applied to discard non-physiological patients. This is described next.

\subsection{Post simulation filter}
\label{section_filter}
Through random sampling, there is a chance that VPs are assigned combinations of arterial network parameters that result in physiologically unrealistic pressures and flow-rate profiles. Thus, to remove these VPs from the VPD, a post simulation filter is applied. ``Hard filters'' are applied to VPs, \textit{i.e.} ranges within which pressure profiles must fall are directly imposed. 
Based on literature reported ranges \cite{sonesson2003intra}, a more conservative version is adopted to allow for the full range of possible pressure waveforms to be expressed in the VPD.
The three conditions of the  the post simulation filter are:
\begin{equation*}
\text{Filters: }
\begin{cases}
\text{max}(\bm{P}_{\text{inlet}}) < 225\text{mmHg}\\
\text{min}(\bm{P}_{\text{inlet}}) > 25\text{mmHg}\\
\text{max}(\bm{P}_{\text{inlet}})-\text{min}(\bm{P}_{\text{inlet}}) <120\text{mmHg}
\end{cases}
\end{equation*}
where $\bm{P}_{\text{inlet}}$ represents the vector describing the time domain pressure profile at the inlet of the system. Using the VPD created through the methodology described above, the ability of ML classifiers to distinguish between healthy and unhealthy VPs is assessed, as outlined next.

\subsection{Representation of measurements}
The output of the pulse wave propagation model is the pressure and flow-rate at all temporal and spatial locations. While, these vectors of pressure or flow-rate at any spatial location (for example $\mathbf{p} = [p(t_0), p(t_0+\Delta t),p(t_0+2\Delta t),\cdots,p(t_0 + k\Delta t)]$) can be used directly as a measurement input to the ML algorithms, its dimensionality is quite large. Furthermore, as severity of stenosis increases, resulting in additional nonlinearities in the model, the time step $\Delta t$ for a stable solution can become very small.
 As pressure and flow-rate profiles are periodic it seems natural to represent the time domain haemodynamic profiles using a FS representation. Using this representation allows the pressure and flow-rate profiles to be described to a high level of completeness in significantly fewer dimensions. Since the input FS coefficients differ by several orders of magnitude between measurements and VPs, they are individually transformed to have zero-mean and unit-variance through the widely used Z-score standardisation \cite{mohamad2013standardization}. The transformed inputs are subsequently used as inputs to the ML algorithms.

With the above stated method of network parameterisation, its sampling, and measurement representation, the ML setup is described next.

\section{Machine learning setup}
\subsection{Test/train split}
The VPD is split into two parts: testing set and training set. The training set is used for learning in the ML algorithms and is set to two-thirds of the size of the VPD. The remaining one-third of the VPs comprise the testing set, which is used to assess the accuracy of the ML algorithms on previously unseen data, i.e. the data not used while training. The ML algorithms are briefly described next.

\subsection{Machine learning algorithms}
\label{sec_ML_setup}
A model mapping a vector of input measurements, $\bm{x}$, to a discrete output classification, $y$, can be described as:
\begin{equation}
y = m(\bm{x}) \quad y \in \bm{\mathcal{C}},
\label{eq_direct_model}
\end{equation}
with,
\begin{equation}
\bm{\mathcal{C}} = \{\mathcal{C}^{(1)}, \mathcal{C}^{(2)},..,\mathcal{C}^{(j)}\},
\end{equation}
where $\bm{\mathcal{C}}$ represents the set describing all possible classifications, and $\mathcal{C}^{(j)}$ represents the $j^{\text{th}}$ possible classification. In the context of this study, the measured inputs  $\bm{x}$ and output classification $y$ represent the haemodynamics measurements taken from VPs and the corresponding health of those VPs, respectively.
The following four ML methods are used in this study.
\subsubsection{Logistic regression (LR)}
The LR classifier \cite{sperandei2014understanding, hilbe2009logistic} is a probabilistic binary classification method. Given that patients belong to one of the two classifications, \textit{i.e.} $\bm{\mathcal{C}} = \{\mathcal{C}^{(1)}, \mathcal{C}^{(2)}\}$, the true binary responses $\tau_i$  are assigned to all subjects in the training set:  
\begin{equation}
  \tau_i =
  \begin{cases}
    1 & \text{if} \quad y_i=\mathcal{C}^{(1)}\\
    0 & \text{if} \quad y_i=\mathcal{C}^{(2)}\\
  \end{cases}.
\label{eq_LR_binary_labels}
\end{equation}
To predict the binary health of a patient an activation function is used. A general equation for an activation function $\text{h}(\bm{x}_{i}, \bm{\theta})$ can be written as:
\begin{equation}
\text{p}\left(\hat{\tau}_i=1 \mid \bm{x}_{i}, \bm{\theta}\right)=\text{h}(\bm{x}_{i}, \bm{\theta}),
\label{eq_hypothesis_general}
\end{equation}
where $\text{p}\left(\hat{\tau}_i=1 \mid \bm{x}_{i}, \bm{\theta}\right)$ represents the predicted probability that the $i^{\text{th}}$ VP belongs to $\mathcal{C}^{(1)}$, given that the patient specific input measurements $\bm{x}_{i}$ have been observed, and that the vector of measurement specific weightings are described by $\bm{\theta}$. Typical choices for $\text{h}(\bm{x}_{i}, \bm{\theta})$ are the the sigmoid and tanh functions. The sigmoid function is shown below:
\begin{equation}
\text{h}(\bm{x}_{i}, \bm{\theta}) = \frac{1}{1 + \exp{(-\bm{\theta}^T \bm{x}_i)}}.
\label{eq_hypothesis_general}
\end{equation}
To obtain optimal measurement specific weightings $\bm{\theta}$, the logistic regression algorithm is trained by minimising the mean error between the predicted probability of VPs producing a positive binary response and the known correct classification across the training set, \textit{i.e.}:

\begin{equation}
\hat{\bm{\theta}} = \argmin_{\bm\theta} \left\{ L(\bm{\theta}, \bm{X}^{\text{train}}, \bm{\tau}^{\text{train}}) \right\}
\label{eq_cost_argmin}
\end{equation}
with
\begin{equation}
\begin{split}
L(\bm{\theta}, \bm{X}^{\text{train}}, \bm{\tau}^{\text{train}}) &=  
-\frac{1}{m}\sum_{i=1}^{m} \Big(  \tau_{i} \log\left(\text{h}(\bm{x}_{i}, \bm{\theta})\right)
+ (1-\tau_{i}) \log\left(1-\text{h}(\bm{x}_{i}, \bm{\theta})\right) \Big),
\end{split}
\label{eq_cost}
\end{equation}
%
%
where $L(\bm{\theta}, \bm{X}^{\text{train}}, \bm{\tau}^{\text{train}})$ represents the average cost, in this case computed as a log loss, across the training set; $\bm{X}^{\text{train}}$ and $\bm{\tau}^{\text{train}}$ represent the matrix of input measurements and the vector of the known correct binary classifications for all the $m$ VPs in the training set, respectively; $\bm{x}_i$ and $\tau_{i}$ represents the vector of input measurements and the known correct binary classification corresponding to the $i^{\text{th}}$ VP, respectively; and $\bm{\theta}$ represents the measurement specific weightings.

The numerical minimisation can be carried out using many algorithms such as gradient descent, gradient descent with momentum \cite{qian1999momentum}, Nesterov accelerated gradient (NAG) \cite{nesterov1983method}, Adadelta \cite{zeiler2012adadelta}, and Adam method \cite{kingma2014adam}. 
Post training, the obtained weightings can be used to predict the health classification of new unseen VPs, \textit{i.e.} VPs within the test set, by equation \eqref{eq_hypothesis_general} through application of a threshold $\mathcal{B}$, often referred to as the decision boundary, to the predicted probabilities as follows:
%
\begin{equation}
  \hat{y}_i =
  \begin{cases}
    \mathcal{C}^{(1)} & \text{if} \quad \text{p}\left(\hat{\tau}_i=1 \mid \bm{x}_{i}, \bm{\theta}\right)\geq \mathcal{B}\\
    \mathcal{C}^{(2)} & \text{otherwise},
  \end{cases}
\label{eq_classify_binary}
\end{equation}
where $\hat{y}_i$ represents the predicted health classification of the new unseen test VP, $\text{p}\left(\hat{\tau}_i=1 \mid \bm{x}_{i}, \bm{\theta}\right)$ represents the predicted probability returned by the activation function through equation \eqref{eq_hypothesis_general}, and $\mathcal{B}$ represents a chosen decision boundary.

The remaining three methods are not described in great detail here. Their descriptions can be found in the references below. LR is described in more detail above as it is later modified for the application in this study.
\subsubsection{Naive Bayes (NB)}
An NB classifier \cite{rish2001empirical, rish2001analysis} is a probabilistic multiclass method. An NB classifier creates a conditional probability model, through the use of Bayes theorem, that predicts the probability of a VP belonging to each  classification, given the measured pressure and flow-rate profiles.

\subsubsection{Support vector machine (SVM)}
An SVM classifier is a non-probabilistic binary classification method \cite{kecman2005support}. An SVM method finds an optimum partition between positive and negative binary outcomes through a high order feature space by maximising the distance between the partition and the nearest instances of both binary outcomes. It is common for SVM classifiers to map the input measurements  to a higher order feature space, typically through the use of an input kernel.

\subsubsection{Random forest (RF) classification method}
An RF classification method is a  non-probabilistic multiclass classification method \cite{liaw2002classification, breiman2001random}. An RF method is an ensemble method, combining the predictions returned by a series of weak decision tree classifiers through the use of a bootstrap aggregation method. Each decision tree within the ensemble is created by repeatedly splitting the training data into subsets, based on an evaluation criteria, to maximise the homogeneity of the subsets.\\

\begin{table*}
\begin{center}
\def\arraystretch{1.2}
\begin{tabular}{| c | c |c |}
\hline
& \textbf{Capable of linear partitions} & \textbf{Capable of non-linear partitions}\\
\hline
\textbf{Probabilistic} & Logistic Regression (LR) & Naive Bayes (NB) \\
\hline
\multirow{2}{*}{\textbf{Non-probabilistic}} & SVM with linear kernel & SVM with radial basis function kernel \\
& & Random Forest (RF) \\
\hline 

\end{tabular}
\caption{The four major classifier behaviour characteristics, and how each classification method aligns with these characteristics.}
\label{table_classifier_characteristics}
\end{center}
\end{table*}

\subsubsection{Motivation for the chosen ML classifers}
Two characteristics that can be used to distinguish between different ML methods are if they are capable of producing linear or non-linear partitions between different classifications, and if they return a probabilistic or non-probabilistic output prediction. These four ML methods are chosen as they encompass all four combinations of classifier characteristic behaviours, as shown in Table \ref{table_classifier_characteristics}. Another attractive feature of these methods is that they all require very little problem specific optimisation. Before ML classifiers are trained and tested using these four different methods, preliminary tests are carried out using the LR method. LR is used for these initial tests as it is computationally inexpensive. Once an initial understanding of the VPD has been gained further classifiers are trained using the other three ML classification methods. The methodologies and considerations required to use the VPD to train and test ML classifiers are explained next.

\subsection{Required size of the VPD}
An important consideration in the creation of VPD is its size---how many virtual patients are sufficient for the ML algorithms to be applied? Here \emph{a priori} evaluation of the required size of the VPD is presented, while \emph{a posteriori} analysis is found in Section \ref{section_PVC}.
A common rule of thumb in ML is that to train a classifier at least 10 examples of each possible classification are required per input dimension, known as events per variable or EPV \cite{vittinghoff2007relaxing}. 
While pressure and flow-rate measurements can be obtained at any location within the arterial network, measurements are limited to the inlet and two outlets of the system, shown in Figure \ref{figure_model_overview} by $P_1$, $Q_1$, $P_2$, $Q_2$, $P_3$, and $Q_3$ respectively. This results in the maximum number of input dimensions to be 66 (each measurement is described by 11 FS coefficients and all six measurements taken). 
A minimum EPV of any one health classification is chosen to be 12 in this study, in order to be on the conservative side of the rule of 10. Two thirds of VPs within the VPD are used for training the classifiers, and the remaining one third are used for testing. From this, it is calculated that the VPD requires 1,188 ($3/2\times12\times66$) VPs with disease in each of the three vessels. Since a balanced data set is desired, the number of healthy patients  required are  3,564 ($1118\times3$). This results in the EPV of 36 for healthy subjects.
 

\subsection{Classifier configurations}
\label{sec_ML_config}

The objectives and configurations of classifiers can be split into two general categories. These two categories are binary classifiers and multiclass classifiers. Binary ML classifiers are trained to predict the outcome of Equation \eqref{eq_direct_model} when the output classification may belong to one of two possible outcomes, \textit{i.e.} $\bm{\mathcal{C}} = \{\mathcal{C}^{(1)}, \mathcal{C}^{(2)}\}$. In contrast, when more than two classes are present, multiclass classifiers are necessary. 

\subsubsection{Binary classifiers}
Binary classifiers are created using one of two different configurations.

\paragraph{Individual vessel binary configuration}
\label{sec_IVBC}

The first configuration of binary classifiers are individual vessel binary classifiers (IVBCs). The purpose of IVBCs are to predict if there is a stenosis present within a particular vessel of a VP's arterial network. When creating IVBCs an arterial vessel of interest must be isolated, and VPs with disease present within this vessel are assigned to the first discrete output classification, $\mathcal{C}^{(1)}$. All other VPs are assigned to the second discrete output classification, $\mathcal{C}^{(2)}$.  The assignment of true state classifications to VPs when creating IVBCs is described by:\\
 \begin{equation}
  {y}_i =
  \begin{cases}
    \mathcal{C}^{(1)} & \text{if disease is present within}
     \text{vessel $a$}\\
    \mathcal{C}^{(2)} & \text{otherwise,}\\
  \end{cases}
\label{eq_ENBC_labels}
\end{equation}
where ${y}_i$ represents the true state classification of the $i^{\text{th}}$ VP, and $a$ represents the arterial vessel for which the binary health is being predicted.

\paragraph{Entire network binary configuration}
The second configuration of binary classifiers are entire network binary classifiers (ENBCs). The purpose of ENBCs is to predict the health of a VP's entire arterial network, \textit{i.e.} irrespective of the vessel in which the disease is located. When creating ENBCs VPs with no disease present within their arterial network are assigned to the first class, $\mathcal{C}^{(1)}$, while all other VPs are assigned to the second discrete output classification, $\mathcal{C}^{(2)}$. The assignment of true state classifications to VPs when creating ENBCs is described by:\\
 \begin{equation}
  {y}_i =
  \begin{cases}
    \mathcal{C}^{(1)} & \text{if no disease is present,}\\
    \mathcal{C}^{(2)} & \text{otherwise.}\\
  \end{cases}
\label{eq_ENBC_labels}
\end{equation}
 Multiclass ML classifiers are discussed next.

\subsubsection{Multiclass classifiers}
\label{sec_multiclass_config}


 Multiclass classifiers predict the outcome of Equation \eqref{eq_direct_model} when the output may belong to more than two different classifications. The purpose of multiclass classifiers is to predict if there is a stenosis present within a VP's arterial network, and if so which vessel does that disease occur within. Thus four different classifications exist:
 \begin{equation}
  \bm{\mathcal{C}} = \{\mathcal{C}^{(1)}, \mathcal{C}^{(2)}, \mathcal{C}^{(3)}, \mathcal{C}^{(4)}\},
  \end{equation}
where $\mathcal{C}^{(1)}$, $\mathcal{C}^{(2)}$, $\mathcal{C}^{(3)}$, and $\mathcal{C}^{(4)}$ represents no disease present; and disease present within the aorta, the first iliac, and  the second iliac respectively. It is found through analysis of binary classification behaviours  (Section \ref{section_binary_analysis}) that LR and SVM classifiers consistently achieve higher accuracy classification than NB and RF classifiers. Thus,  multiclass classifiers are only created using these two methods. However, LR and SVM methods are both inherently binary---only naturally capable of distinguishing between two classes. In order to be used as multiclass classifiers they can be adopted through strategies such as one-versus-all  \cite{rifkin2004defense} and one-versus-one  \cite{rocha2013multiclass}. These are described next.


\paragraph{One-versus-all (OVA)}
\label{sec_ova_config}

 An OVA strategy  \cite{rifkin2004defense} trains multiple instances of binary classifiers, each designed to predict the probability of a separate classification problem. These probabilities are then combined to make a multiclass prediction.
 
 In our problem, the OVA strategy trains four instances of a binary classifiers. Each binary classifier prescribes a correct binary health classification of $1$ to all VPs belonging to the corresponding possible classification. All other patients, irrespective of which of the other three classifications they belong too, are assigned a correct binary health classification of $0$:
\begin{equation}
  \tau^{(j)}_i =
  \begin{cases}
    1 & \text{if} \quad y_i = \mathcal{C}^{(j)}\\
    0 & \text{otherwise}
  \end{cases}, \quad j \in \{1,2,3,4\},
\end{equation}
where $\tau^{(j)}_i$ represents the correct binary health classification of the $i^{\text{th}}$ VP for the $j^{\text{th}}$ instance of a binary classifier. 
To assign a predicted multiclass classification to a new subject, the predicted probability of producing a positive binary response ($y_i = \mathcal{C}^{(j)}$) is found for all the four binary classifiers. The classification that corresponds to the highest predicted probability is then selected as the multiclass prediction.\\

\paragraph{One-versus-one (OVO)}
\label{sec_ovo_config}

An OVO strategy  \cite{rocha2013multiclass} creates binary classifiers for all the pairs of the classes. Thus if $n$ total classes exist, then $n(n-1)/2$ binary classifiers are created. The most frequent class predicted among these binary classifiers is then used as the multiclass prediction.


In our problem, the OVO strategy creates six instances of a binary classifier. Each binary classifier is designed to distinguish between two different classes.  Thus,  the binary classifier  created to distinguish between classifications $\mathcal{C}^{(j)}$ and $\mathcal{C}^{(k)}$ uses:
\begin{equation}
  \tau^{(j,k)}_i \!=\!
  \begin{cases}
    1 & \text{if} \quad y_i = \mathcal{C}^{(j)}\\
    0 & \text{if} \quad y_i = \mathcal{C}^{(k)}
  \end{cases}, \;  j,k \in \{ 1,2,3,4\} , j \!\neq\!k.
\end{equation}
%
When predicting the classification of an unseen test VP, a voting scheme is applied. The input measurements taken from the test VPs are passed through each of the six instances of a binary classifier, and the predicted classifications recorded. The  classification that occurs most frequently is selected as the multiclass prediction.

It is found that while both LR classifiers employing an OVA method and SVM classifiers employing an OVO method achieve high aortic, first iliac, and second iliac classification accuracy, they produce very low healthy VP classification accuracy (see Section \ref{section_multiclass_analysis}). To rectify the low healthy VP classification accuracies a custom probabilistic configuration is developed, as described next. 

\paragraph{Custom probabilistic configuration (CPC)}
\label{sec_cpc_config}

The CPC method assigns all VPs a health classification corresponding to `no disease' before running any binary classifiers. This strategy treats `no disease present' as the opposite to the three other possible classifications a VP may belong to. 
The binary classifiers employed in CPC are identical to OVA, except that the classifier for `no disease' is omitted. Thus, as opposed to four binary classifiers in the OVA strategy, this strategy uses only three binary classifers---each pertaining to diseased aorta, first iliac, and second iliac, respectively.
The assignment of true state binary outcomes to VPs for the three binary classifiers are:
\begin{equation}
  \tau^{(j)}_i =
  \begin{cases}
    1 & \text{if} \quad y_i = \mathcal{C}^{(j)}\\
    0 & \text{otherwise}.
  \end{cases}, \quad j \in \{2,3,4\}.
\end{equation}
%
Note that $j=1$ for `no disease' classification is not included. To predict a muliclass classification for test VPs, the vessel that produces the highest probability of being diseased among the three binary classifiers is first found. The default multiclass classification is `no disease' unless the highest probability of disease occurring is greater than a prescribed threshold (decision boundary), in which case the test VP is predicted to have disease in the arterial vessel with this highest probability, \textit{i.e.} 
 \begin{equation}
 \small
 \label{eq_CPC}
  \hat{y}_i =
  \begin{cases}
    \mathcal{C}^{(1)} & \text{if max}\left(\text{p}\left(\hat{\tau}^{(j)}_i=1 \mid \bm{x}_{i}, \bm{\theta}^{(j)}\right)\right) 
    < \mathcal{B} \text{ for }j\in \{2, 3, 4\},\\[10pt]
    
   \mathcal{C}^{(2)} & \text{if p}\left(\hat{\tau}^{(2)}_i=1 \mid \bm{x}_{i}, \bm{\theta}^{(2)}\right) = 
   \text{max}\left(\text{p}\left(\hat{\tau}^{(j)}_i=1 \mid \bm{x}_{i}, \bm{\theta}^{(j)}\right)\right)
   \text{for }j\in \{2, 3, 4\}
    \text{ and p}\left(\hat{\tau}^{(2)}_i=1 \mid \bm{x}_{i}, \bm{\theta}^{(2)}\right) \geq \mathcal{B} ,\\
   ...\\
   \mathcal{C}^{(4)} & \text{if p}\left(\hat{\tau}^{(4)}_i=1 \mid \bm{x}_{i}, \bm{\theta}^{(4)}\right) = 
   \text{max}\left(\text{p}\left(\hat{\tau}^{(j)}_i=1 \mid \bm{x}_{i}, \bm{\theta}^{(j)}\right)\right)
     \text{for }j\in \{2, 3, 4\}
   \text{ and p}\left(\hat{\tau}^{(4)}_i=1 \mid \bm{x}_{i}, \bm{\theta}^{(4)}\right) \geq \mathcal{B},
  \end{cases}
\end{equation}
where $\text{p}\left(\hat{\tau}^{(j)}_i=1 \mid \bm{x}_{i}, \bm{\theta}^{(j)}\right)$ represents the probability of the $i^{\text{th}}$ VP being predicted to produce a positive binary response for the $j^{\text{th}}$ instance of a classifier within the ensemble; $\bm{x}_{i}$ represent the vector of measurements for the test patient, $\bm{\theta}^{(j)}$  represent the measurement specific weightings for the $j^{\mathrm{th}}$ classifier; and $\mathcal{B}$ represents the threshold (decision boundary).


As opposed to the classical OVA, where the classification with highest predicted probability, irrespective of the magnitude of this probability, is chosen, CPC requires a minimum certainty of disease being present to be met before the default hypothesis `no disease' can be overridden. It is not possible to create multiclass classifiers in this manner using non-probabilistic methods, such as SVM.

\subsection{Quantification of results}

\begin{figure}[tb]
\centering
\includegraphics[width=2.5in]{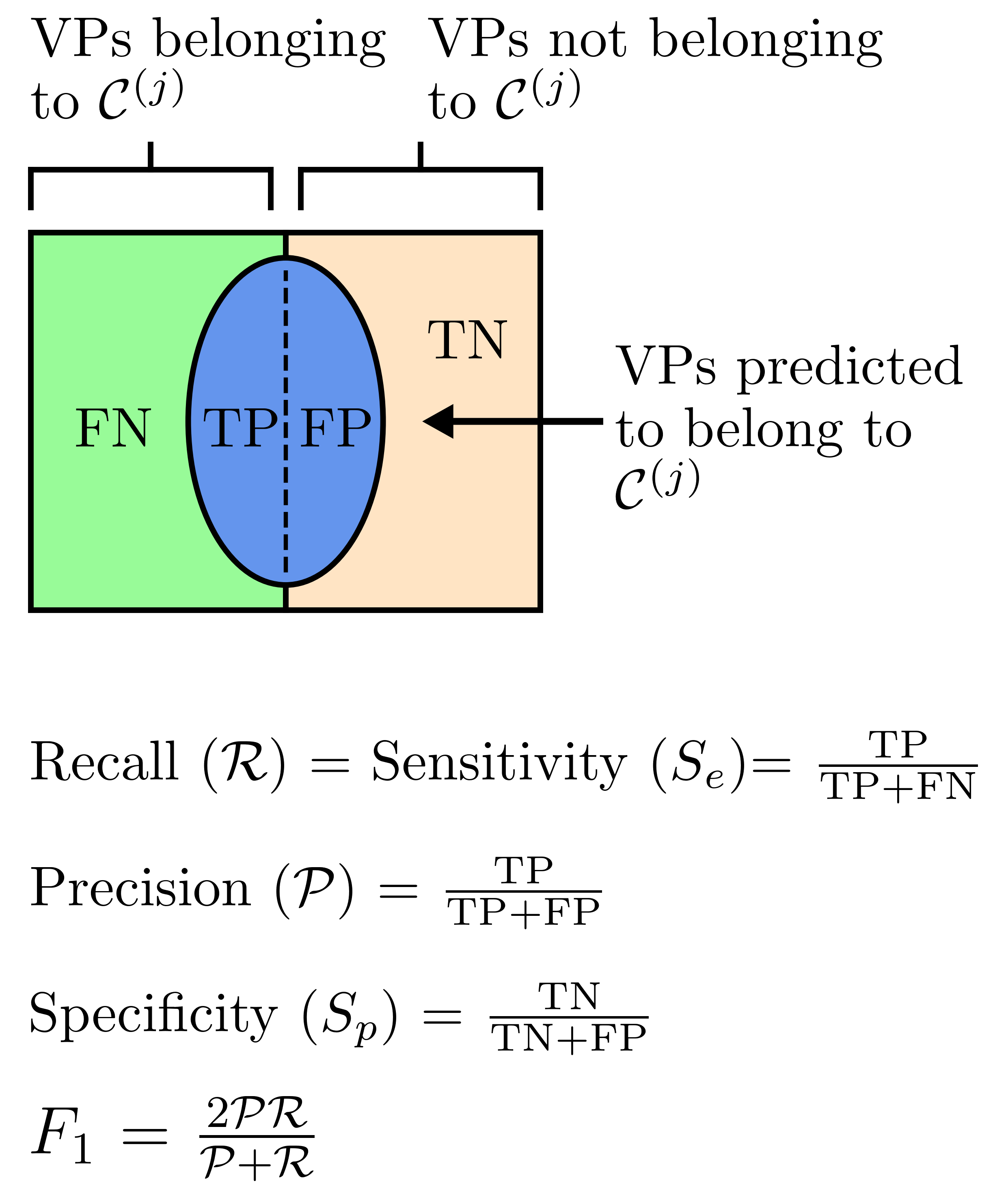}
\caption{Computation of sensitivity, specificity, recall, and precision are shown above. TP: True Positive; FN: False Negative; FP: False positive; and TN: True Negative.}
\label{figure_precision_recall}
\end{figure}

Two different methods are used to quantify and compare the results of different classifiers. The first, also the most intuitive, of these is to compute the sensitivity and specificity of classification across the test set. Determination of whether a VP is classified correctly or incorrectly can be achieved by comparison against the true states, see Table \ref{table_TP_FP}. The proportion of VPs belonging to a classification that are correctly classified, \textit{i.e.} the sensitivity ($S_e$), is computed using the equation $S_e$ =TP/(TP+FN), while the proportion of VPs not belonging to a classification that are correctly classified, \textit{i.e.} the specificity ($S_p$), is compute using the equation $S_p$=TN/(TN+FP). The relationships between the TP, FN, FP, TN, $S_e$, and $S_p$ with respect to the class $\mathcal{C}^{(j)}$ are shown in Figure \ref{figure_precision_recall}.

\begin{table*}
\begin{center}
\def\arraystretch{1.2}
\begin{tabular}{| c | c |c |}
\hline
& \textbf{VP belongs to}  & \textbf{VP does not belongs to}\\
& \textbf{classification $\bm{\mathcal{C}^{(j)}}$} &  \textbf{classification $\bm{\mathcal{C}^{(j)}}$} \\
\hline
\textbf{VP predicted to belong} & \multirow{2}{*}{$\mathcal{C}^{(j)}$ True positive (TP)} & \multirow{2}{*}{$\mathcal{C}^{(j)}$ False positive (FP)} \\
\textbf{to classification $\bm{\mathcal{C}^{(j)}}$} & & \\
\hline
\textbf{VP predicted to not belong} &  \multirow{2}{*}{$\mathcal{C}^{(j)}$ False negative (FN)} & \multirow{2}{*}{$\mathcal{C}^{(j)}$ True negative (TN)} \\
\textbf{to classification $\bm{\mathcal{C}^{(j)}}$} & & \\
\hline 
\end{tabular}
\caption{Definitions of true/false positives and true/false negatives for a particular class $\mathcal{C}^{(j)}$.}
\label{table_TP_FP}
\end{center}
\end{table*} 

In the case of multiclass classifiers, assessment of the accuracy of classification requires provision of the sensitivity and specificity corresponding to each classification. In our case, there are four classes, thus requiring specification of eight different numbers (four sensitivities and four specificities). While quantifying the accuracy of ML classifiers through the sensitivity and specificity of each classification is simple and easily understood, the description of results through two different numbers per classification can make comparison of different classifiers difficult.

 A more complex, however easier to compare, method for quantifying the accuracy of ML classifiers is the $F$ score \cite{SasakiThe}. The $F$ score produces a single quantitative score allowing for easy comparison. Higher values of $F$ score imply a better classification. To calculate the $F$ score the precision ($\mathcal{P}$) and recall ($\mathcal{R}$) of each discrete classification are calculated. A visual explanation of the F score, precision, and recall is shown in Figure \ref{figure_precision_recall}. Precision is the proportion of patients predicted to belong to a classification, who do in fact belong to that classification. The recall is the portion of patients belonging to a classification who are correctly classified, thus identical to sensitivity. The $F$ score combines the precision and recall as follows:
\begin{equation}
F=\frac{({\delta}^2+1)\mathcal{PR}}{\delta^2\mathcal{P}+\mathcal{R}},
\label{eq_F1}
\end{equation}
where $\mathcal{P}$ represents the precision, $\mathcal{R}$ represents the recall, and $\delta$ represents a hyper parameter. Values of $\delta$ above 1 give preference to recall, while values under 1 give preference to precision. Although there is a preference to recall in the proposed application of the classifiers, $\delta=1$ is used to get a general sense of classifier performance without any bias. As $\delta=1$ is  used, the $F$ score is referred to as the $F_1$ score and is, thus, essentially the harmonic mean of precision and recall.

While the $F_1$ score balances the affect of precision and recall, it does not balance the affect of the sensitivity and specificity. Given a situation in which there is an equal number of healthy and unhealthy VPs, an ENBC which correctly predicts the health of 80\% of healthy VPs ($\mathcal{R}=S_e=0.8$) and 20\% of unhealthy VPs ($S_p=0.2$) will achieve an $F_1$ score of 0.61. An ENBC that correctly predicts the health of 20\% of healthy VPs ($\mathcal{R}=S_e=0.2$) and 80\% of unhealthy VPs ($S_p=0.8$), however, will achieve an $F_1$ score of 0.28, despite the fact that the total number of VPs who have been correctly classified is unchanged. This highlights the importance of using both the $F_1$ score and the sensitivities/specificities in combination.
\section{Results and discussion}

\subsection{Empirical evaluation VPD size}
\label{section_PVC}
While an estimation to the adequacy of the VPD size has been made by calculating the EPV, this can be checked more thoroughly by training and testing a series of classifiers with successively increasing number of VPs. This assessment is made for the case with the largest input dimensionality, i.e. when all the six measurements---three pressure and three flow-rate profiles---at all the three  measurement locations are used (see Figure \ref{figure_model_overview}).

To minimise the lowest number of VPs belonging to a single classification, classifiers must be trained to predict the health of each vessel individual. As the VPD has been created so that there is an equal number of healthy and unhealthy VPs, for any given number of available VPs an ENBC will have half of the number of available VPs belonging to $\mathcal{C}^{(1)}$ and half belonging to $\mathcal{C}^{(2)}$. On the contrary, three series of IVBCs are created (as described in Section \ref{sec_IVBC}), each predicting the health of a different vessel. This results in each instance of an IVBC having 5/6 of the available VPs belonging to a negative binary classification, however only 1/6 of the number of available VPs belonging to a positive binary classification. By empirically showing there is an adequate number of VPs to train and test classifiers in this extreme situation, it is reasonable to assume there is an adequate number of VPs to train and test ENBCs.

Due to the class imbalance present, \textit{i.e.} there are significantly more VPs belonging to $\mathcal{C}^{(2)}$ than $\mathcal{C}^{(1)}$, a weighting $w$ is applied to the cost of VPs belonging to $\mathcal{C}^{(1)}$ when training IVBCs. Without this weighting, the IVBCs are biased towards VPs belonging to $\mathcal{C}^{(2)}$. The weighting applied to the cost of the prediction of VPs belonging to $\mathcal{C}^{(1)}$ for each classifier is calculated by assigning a ratio $r$ to the effective number of VPs belonging to classifications $\mathcal{C}^{(1)}$ and $\mathcal{C}^{(2)}$:  
\begin{equation}
r=\frac{w*m^{(1)}}{m^{(2)}},
\label{eq_ratio}
\end{equation}
where 
$m^{(1)}$ and $m^{(2)}$ represent the number of VPs belonging to classes $\mathcal{C}^{(1)}$ and $\mathcal{C}^{(2)}$, respectively. 
The corresponding cost (loss) function is modified from equation \eqref{eq_cost} to include the weight $w$ as
\begin{equation}
\begin{split}
L(\bm{\theta}, \bm{X}^{\text{train}}, \bm{\tau}^{\text{train}}) &=  
-\frac{1}{m}\sum_{i=1}^{m} \Big(  w\; \tau_{i} \log\left(\text{h}(\bm{x}_{i}, \bm{\theta})\right)
+ (1-\tau_{i}) \log\left(1-\text{h}(\bm{x}_{i}, \bm{\theta})\right) \Big),
\end{split}
\label{eq_cost_mod}
\end{equation}
When $r=1$ is used, VPs belonging to $\mathcal{C}^{(1)}$ and $\mathcal{C}^{(2)}$ have the potential to contribute equally to the total cost of prediction across the training set. If $r>1$ is used, bias is given towards VPs belonging to $\mathcal{C}^{(1)}$, and $r<1$ gives bias towards VPs belonging to $\mathcal{C}^{(2)}$. Unless stated otherwise, $r=1$ is used.

\begin{figure}[tb]
\centering
\includegraphics[width=3in]{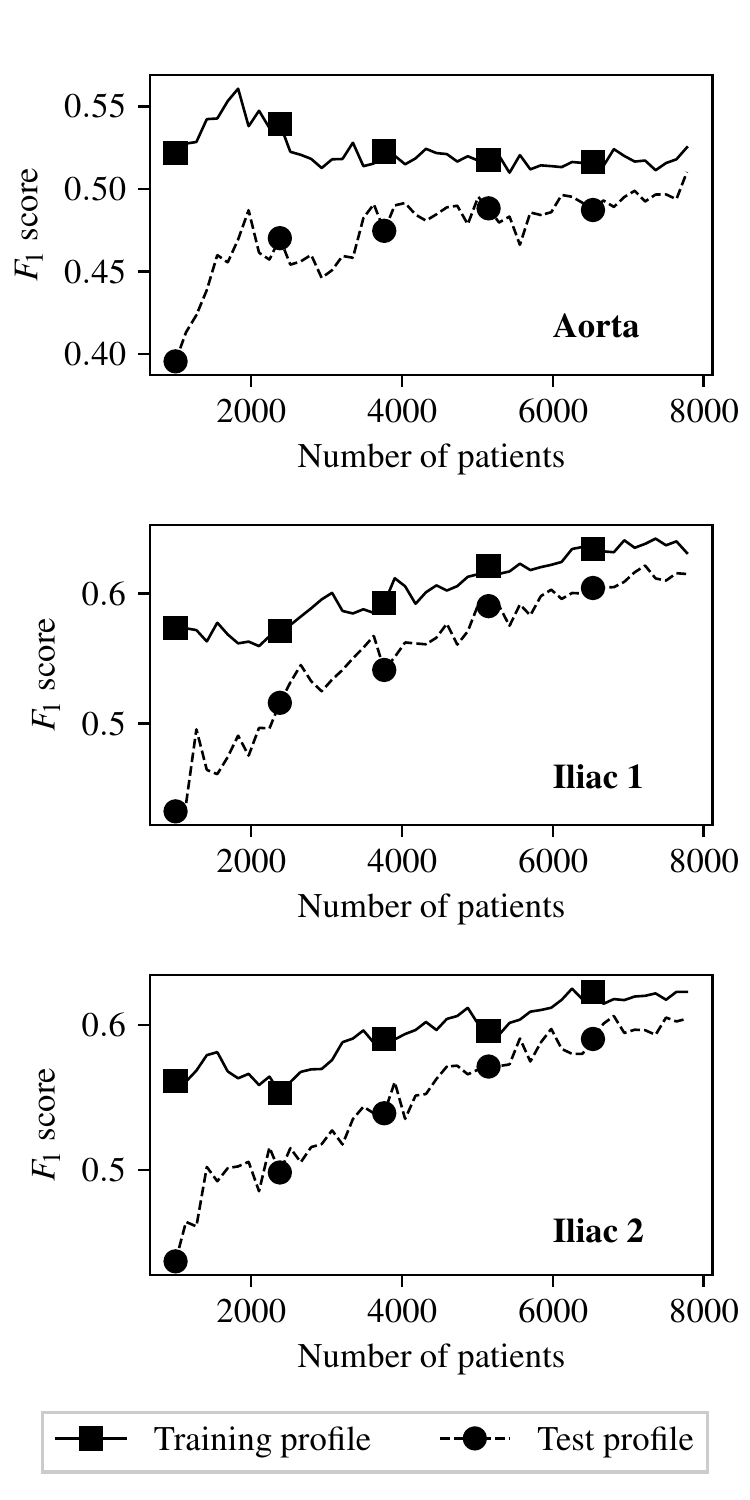}
\caption{Analysis of the adequacy of VPD size when using pressure and flow-rate measurements at all the three locations: training and test $F_1$ scores with successively increasing VPD size.}
\label{fig_patient_number_variance}
\end{figure}

For successively increasing number of VPs, five instances of each of the three IVBCs corresponding to disease in the three vessels are trained and tested. Each of these instances uses a different random subset of VPs for training and testing the classifier. The average performance of these five instances is then computed, thus minimising the effect of test-train split. This is referred to as five fold validation. The average $F_1$ scores achieved across the training and test sets, over the five folds, with increasing numbers of VPs are shown in Figure \ref{fig_patient_number_variance}.

Figure \ref{fig_patient_number_variance} shows that both training and test accuracies are low when a small proportion of the VPD is made available to ML classifiers. This suggests that the classifiers being trained are underfitting the training data, \textit{i.e.} low variance but high bias. The classifiers trained can neither fit the training data nor generalise to the test data. As the number of available VPs increases the behaviour of classifiers differs between the aorta and two common iliacs. In the case of the aorta, the training accuracy remains relatively constant, while the test accuracy increases. In the case of the two common iliac classifiers, both the training and test accuracy increase. These behaviours suggest the classifiers are fitting the training data better, and as a consequence are better able to classify test patients. Initially, between 1,000 and 5,000 available VPs, the changes made to the partition between VPs belonging to $\mathcal{C}^{(1)}$ and $\mathcal{C}^{(2)}$ through the input measurement space are significant, and so there are large jumps in change to the training and test accuracies. As the number of available VPs continuous to increase the partition between healthy and unhealthy patients through the input measurement space begins to converge to an optimum solution. This causes the changes to the training and test accuracies to reduce, and eventually flatten off. Figure \ref{fig_patient_number_variance} suggests that beyond 7,000 VPs the VPD contains enough VPs to train and test ML classifiers. This is shown by the fact that the training and test accuracies of each vessel are consistent for the final several numbers of available VPs, and so the partitions between healthy and unhealthy patients are no longer changing. 

\subsection{ENBC results}
\label{section_binary_analysis}

The architecture of LR, NB, and SVM classifiers can all be considered to be problem independent. While these three algorithms are able to undergo varying levels of problem specific optimisation, the underlying structure of the classifier cannot be changed. In the case of SVM classifiers, the classifier is optimisation for a specific problem by choosing a kernel to map the input measurements to a higher order feature space. Unless otherwise stated, all SVM classifiers use a radial basis function kernel. In the case of NB classifiers, the classifier is optimisation to a specific problem by choosing the distribution of input measurements across the data set. Here, for NB, it is assumed that all input measurements are normally distributed across the data set.

 The architecture of RF classifiers, however, is dependent on the specific problem. The number of trees within the ensemble and the maximum depth of each tree can be optimised for a specific problem. To fit the RF classifiers a basic grid search is carried out. The hyperparameters describing the architecture that produces the highest $F_1$ score is empirically found, and this combination of hyperparameters is then used for all further classifiers trained and tested.
 
\begin{sidewaysfigure}
\centering
\includegraphics[width=9.0in]{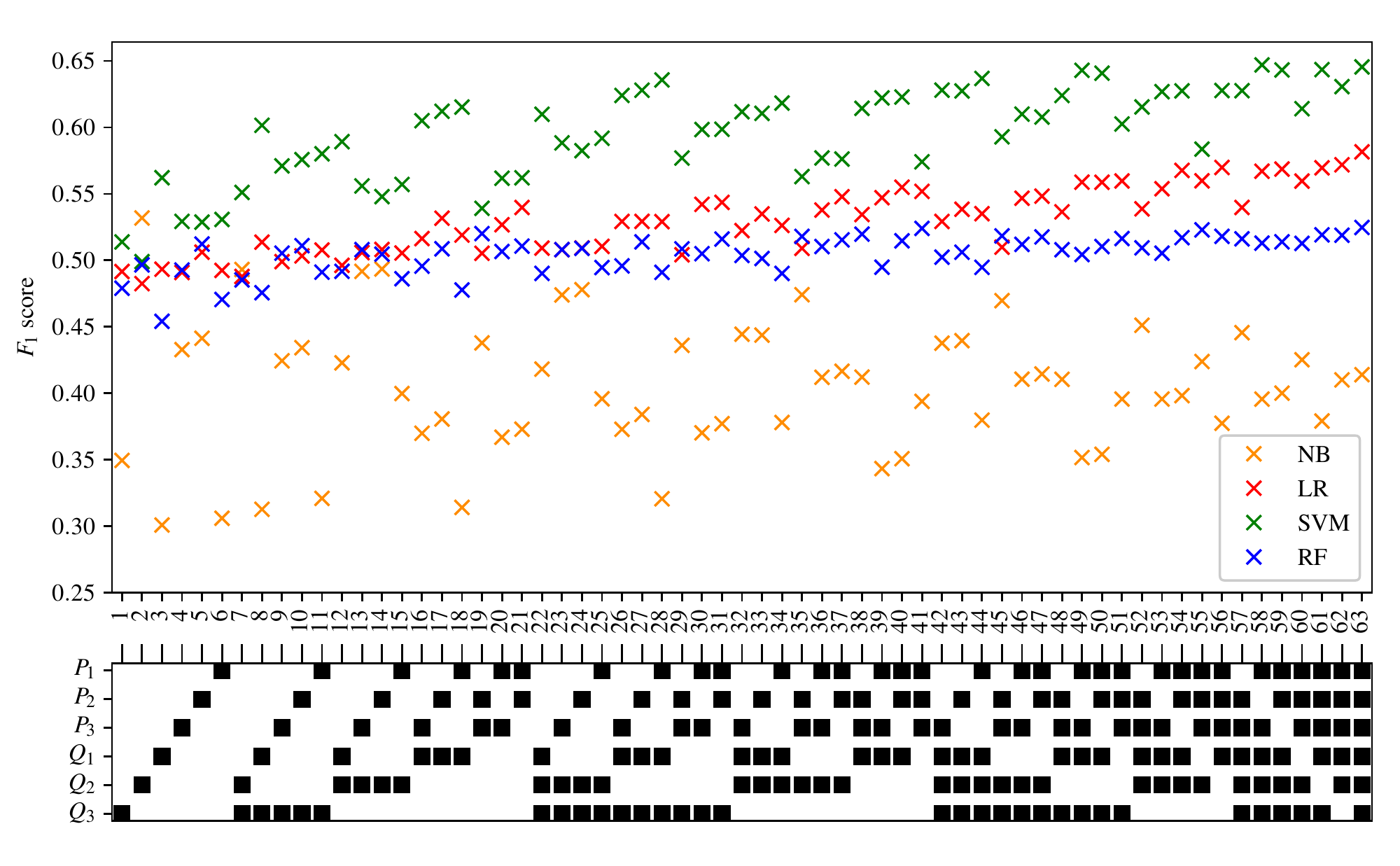}
\caption{The $F_1$ scores achieved by the ENBCs employing the NB, LR, SVM, and RF methods for all the combinations of the input measurements. The bottom legend shows the measurements used in black squares.}
\label{fig_combination_inputs}
\end{sidewaysfigure}

There are 63 possible combinations of input measurements that can be provided to the ML classifiers from the three locations at which pressure and flow-rate are measured. A combination search is performed---for every combination of input measurements, an ENBC is trained and then subsequently tested using each of the four different classification methods. The average $F_1$ score, sensitivity, and specificity of healthy classification accuracy for each input measurement combination and classification method across five folds are recorded. Combinations of interest are then further analysed. The full tables of results are shown in Appendix \ref{appendix_A}. The $F_1$ score achieved by each ML method and combination of input measurements are visually shown in Figure \ref{fig_combination_inputs}. 

\subsubsection{Like-for-like input measurement comparison.}
\begin{figure}[tb]
\centering
\includegraphics[width=2.5in]{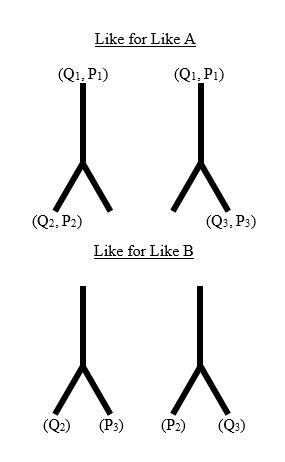}
\caption{Two examples of like for like input measurements.}
\label{fig_like_for_like}
\end{figure}

To gain a better understanding of how much difference in $F_1$ score can be considered insignificant, classifiers that should theoretically achieve identical accuracies are compared. 
Exploiting the symmetrical structure of the arterial network (see Figure \ref{figure_model_overview}), classifiers that use symmetric measurements can be identified. These are referred to as like-for-like measurements; two examples of such measurements are shown in Figure \ref{fig_like_for_like}.
Any discrepancy between the $F_1$ scores achieved by classifiers trained using like-for-like input measurement combinations is therefore introduced due to training and statistical errors. 

\begin{figure}[tb]
\centering
      \includegraphics[width=3.7in]{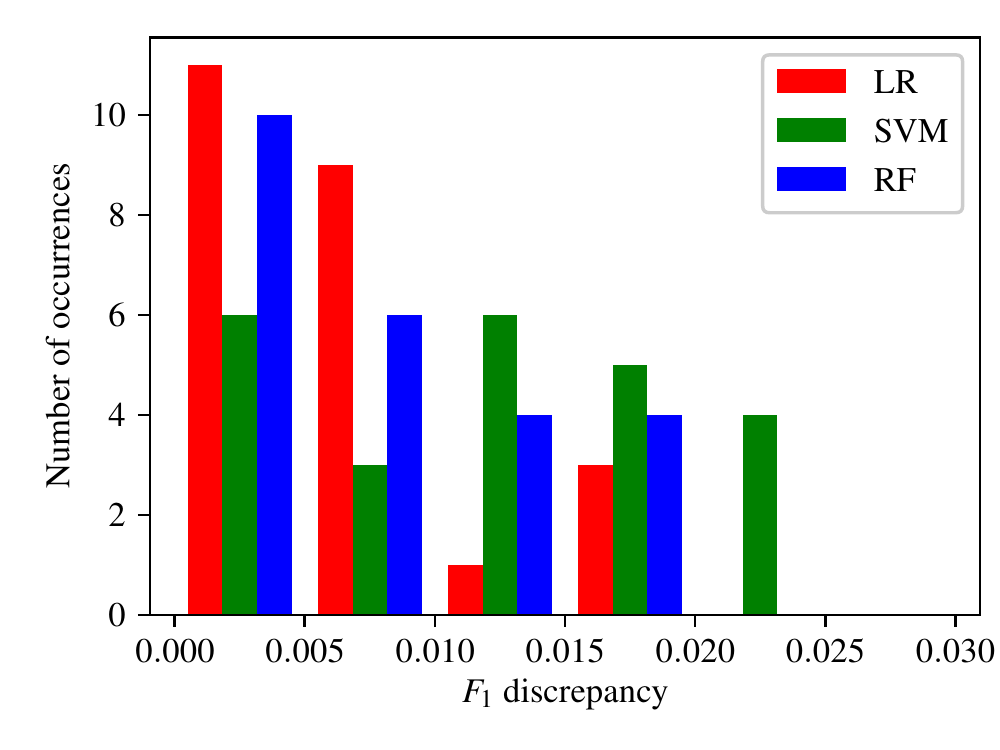}
\caption{Histograms of the discrepancy between the $F_1$ scores of `like for like' ENBCs.}
\label{fig_discrepancy_stenosis}
\end{figure}

There are 24 possible cases of like-for-like input measurement pairs. The discrepancy in the $F_1$ score achieved by the two classifiers within each of these pairs is computed  when using each of the four different classification methods. 
It is found that NB classifiers show significantly greater magnitudes in the discrepancies of $F_1$ scores produced than any of the three other methods. The maximum discrepancy in $F_1$ scores produced when using an NB method is equal to 0.18. This large discrepancy points to something beyond statistical and training errors and is, therefore, most likely related to the unsuitability of the NB method to our problem.
It is therefore decided to exclude the results achieved by the NB method from all subsequent analysis. The histograms of the discrepancies in the $F_1$ score between like for like input measurement combinations produced when using the remaining three ML methods are shown in Figure \ref{fig_discrepancy_stenosis}.

%

%
%
%
%

 \begin{figure}
\centering
     \includegraphics[width=3.7in]{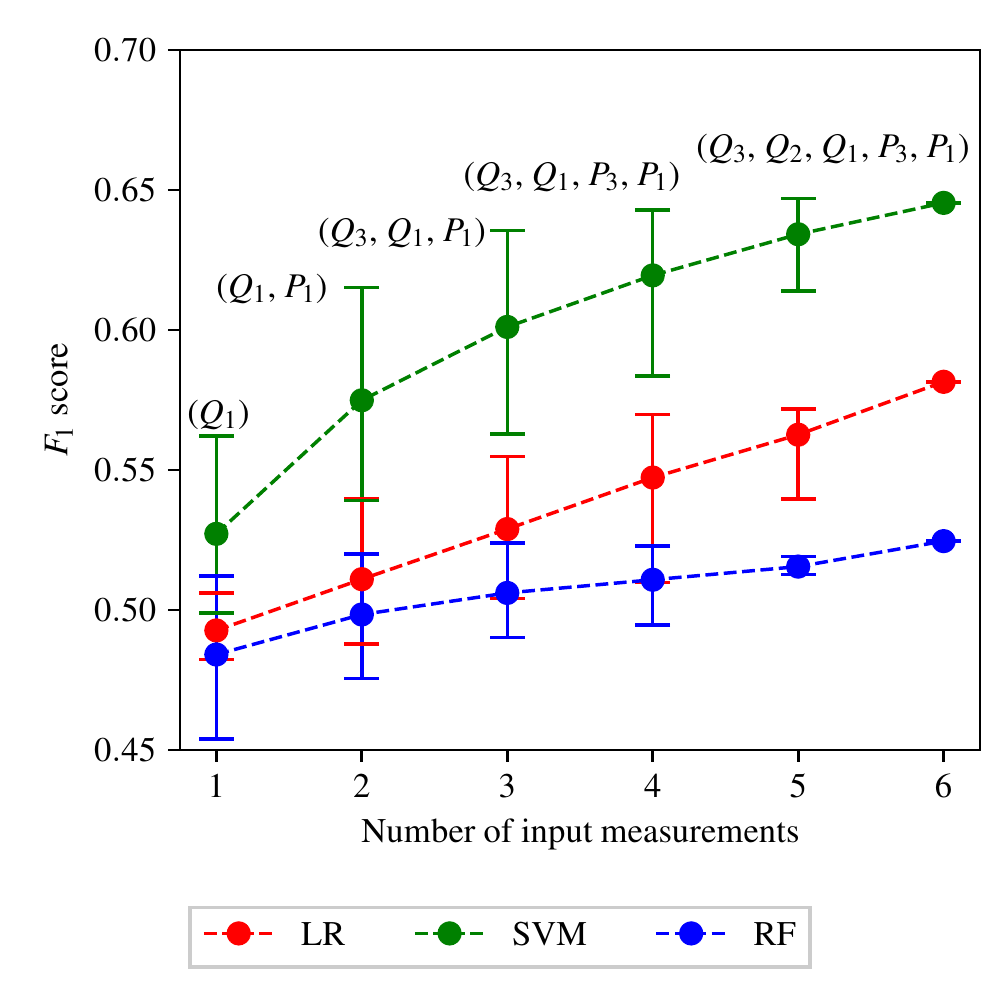}
\caption{The average, maximum, and minimum $F_1$ score achieved by all the ENBCs against the numbers of input measurements. The central markers represent the average score achieved, while the error bars indicate the upper and lower limits. The combination of input measurements that produces the highest $F_1$ score is identified in text for the SVM method.}
\label{fig_number_measurements}
\end{figure}

Figure \ref{fig_discrepancy_stenosis} shows that the discrepancy in $F_1$ scores between like-for-like input measurement combinations follow a very similar pattern for both the LR and RF classification methods. For both of these methods it can be seen that the majority of the 24 like-for-like input measurement combinations produce a discrepancy in $F_1$ score of less than 0.005. There is then a clear exponential decay in the number of occurrences as the $F_1$ score discrepancy increases. 20 of the 24 LR pairs, and 16 of the 24 RF pairs achieved a discrepancy of less than 0.01. When looking at the $F_1$ discrepancies of SVM classifiers, there appears to be no real decay in the number of occurrences as the $F_1$ discrepancy increases, and instead a relatively constant number of SVM pairs produce $F_1$ discrepancies between 0 and 0.025.

The maximum discrepancy in $F_1$ scores between like-for-like input measurement combinations is equal to 0.0231. This discrepancy in $F_1$ score is measured between two pairs of input measurements when using an SVM method. The firsts of these two pairs is ($Q_3$, $P_3$) and ($Q_2$, $P_2$). When training a SVM classifier using ($Q_3$, $P_3$) the sensitivity and specificity is equal to 0.71 and 0.51 respectively. When training an SVM classifier using ($Q_2$, $P_2$) the sensitivity and specificity is equal to 0.74 and 0.47 respectively. The second pair of input measurements producing a discrepancy in $F_1$ score of 0.0231 is ($Q_3$, $P_1$) and ($Q_2$, $P_1$). When training SVM classifiers using ($Q_3$, $P_1$) and ($Q_2$, $P_1$) the sensitivities and specificities are equal to 0.76 and 0.50; and 0.8 and 0.46 respectively. It can be seen that in the case of both pairs of input measurements highlighted above, while there are some differences in the sensitivities and specificities produced, the differences in accuracies are relatively low and the behaviours of each of the two classifiers are relatively consistent.

From Figure \ref{fig_discrepancy_stenosis}  and the aforementioned analysis, a difference in $F_1$ score of more than 0.01 between two LR, SVM, or RF classifiers trained using different input measurements can be considered to be significant and likely due to the behaviour of the classifiers. It is important to remember, however, that a difference in $F_1$ score of approximately 0.025 is required to fully rule out the possibility that patterns are due to training or statistical errors.\\
 
\subsubsection{Effect of the number of input measurements} 
 
 Appendix \ref{appendix_A} and Figure \ref{fig_combination_inputs} show that there is a correlation between the number of input measurements used in the ML classifiers and the $F_1$ score. To investigate this further the average $F_1$ score achieved by  all the classifiers  using one to six input measurements is found for each of the three different classification methods. The maximum and minimum $F_1$ scores are also recorded and shown in Figure \ref{fig_number_measurements}. 
It can be seen that as the number of input measurements increases, the average $F_1$ score achieved by all classification methods also increases. The increase in $F_1$ score is most noticeable for the SVM method. For the LR and RF classification methods, the average $F_1$ score achieved when using 1 input measurement is approximately 0.5, representing naive classification ($S_e+S_p=1$). The average $F_1$ score achieved by SVM classifiers trained using 1 input measurement is marginally better than naive classification. This finding that the average $F_1$ score increases as the number of input measurements increases is expected as the discriminatory information increases, on average, as more measurements are made available.

Observing the range of maximum to minimum $F_1$ scores in Figure \ref{fig_number_measurements} it can be seen that as the number of input measurements increases, the range of $F_1$ scores decreases. An interesting pattern to note is that while the average and minimum $F_1$ score achieved increases when increasing the number of input measurements between four and six, the maximum remains relatively constant. The maximum and minimum $F_1$ scores are shown in Table \ref{table_max_min_svm}, along with the corresponding sensitivities and specificities. 
Table \ref{table_max_min_svm} shows that the maximum accuracy of classifications---assessed by $F_1$ scores, sensitivities, and specificities---vary insignificantly between four, five, and six measurements. Thus, the analysis points that similar levels of accuracies can be achieved by using only four measurements compared to the case when all six measurements are used, but one must be judicious in the choice of the four measurements.

\begin{table*}
\begin{center}
\def\arraystretch{1.2}
\begin{tabular}{ |c | c | c |c  c c |}
\hline
\textbf{Number of input }  & \textbf{Importance} & \textbf{Combination} & \textbf{$F_1$} & \textbf{Sensitivity} &  \textbf{Specificity }\\
\textbf{measurements }&  &  &\textbf{score} &  \textbf{}  &  \textbf{} \\
\hline

\multirow{2}{*}{\textbf{4}} & \textbf{Maximum}  & ($Q_3$, $Q_1$, $P_3$, $P_1$) &0.6429 & 0.7994 & 0.5688\\
 & \textbf{Minimum}  & ($Q_2$, $P_3$, $P_2$, $P_1$) & 0.5836 & 0.8059 & 0.4920 \\
\hline
\multirow{2}{*}{\textbf{5}} & \textbf{Maximum}  & ($Q_3$, $Q_2$, $Q_1$, $P_3$, $P_1$) & 0.6469 & 0.8115 & 0.5683\\
 & \textbf{Minimum}  & ($Q_3$, $Q_2$, $P_3$, $P_2$, $P_1$) & 0.6140 & 0.7947 & 0.5340\\
\hline 
\textbf{6} & - & ($Q_3$, $Q_2$, $Q_1$, $P_3$, $P_2$, $P_1$) & 0.6454 & 0.8050 & 0.5694 \\
\hline
\end{tabular}
\caption{The combinations of input measurements that produce the maximum and minimum $F_1$ scores when providing four, five, and six input measurements. The corresponding sensitivities and specificities are also included.}
\label{table_max_min_svm}
\end{center}
\end{table*} 

\subsubsection{Importance of inlet pressure and flow-split}
\label{section_important_measurements}

\begin{figure}[tb]
\centering
\includegraphics[width=3.7in]{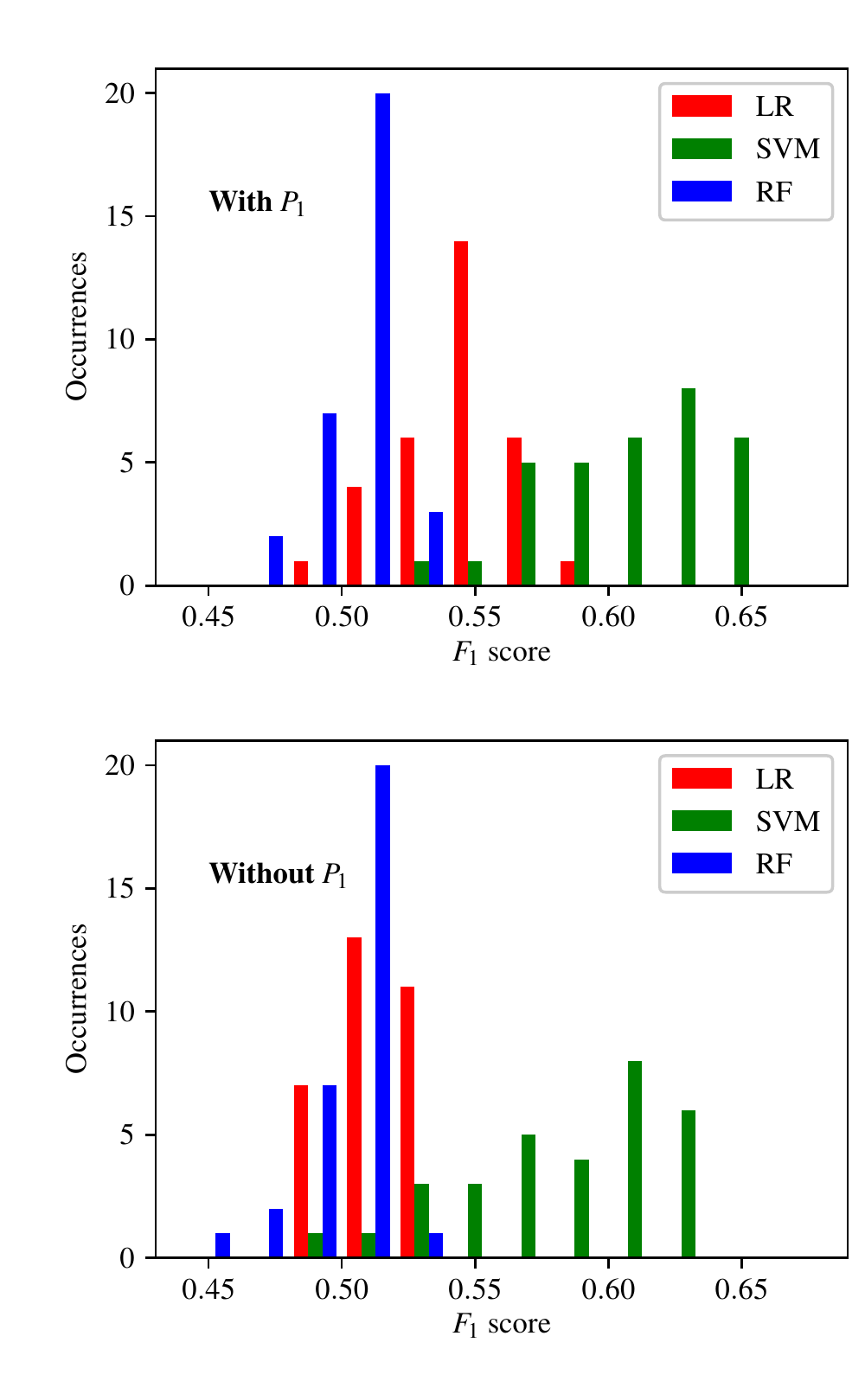}
\caption{The histograms of the $F_1$ scores achieved by the ENBCs that include $P_1$ (upper), and exclude $P_1$ (lower).}
\label{fig_P1_stenosis}
\end{figure}

 A further pattern noticed within the tables in Appendix \ref{appendix_A} and Figure \ref{fig_combination_inputs} is that classifiers trained using $P_1$ generally perform better than those that do not use $P_1$. To analyse this further, the $F_1$ scores of classifiers trained with and without $P_1$ are separated and plotted in Figure \ref{fig_P1_stenosis}. For LR and SVM classifiers, a clear improvement of  $\Delta F_1 \approx 0.05$ is observed when $P_1$ is included. This behaviour is expected, in part due to design. There are a total of 32 combinations of input measurements that include $P_1$, and 31 combinations of input measurements that exclude $P_1$.  The classifier trained using all six input measurements, and five of the six classifiers trained using five input measurements contain $P_1$. Only one classifier trained using five input measurements does not include $P_1$. It has previously been shown in Figure \ref{fig_number_measurements} that, generally, classifiers trained using more input measurements achieve higher accuracy classification results.  There is, therefore, some expected skewing towards higher $F_1$ scores in favour of classifiers trained with $P_1$. This expected behaviour is further amplified by the fact that only one combination of input measurements consists of a single input measurement and contains $P_1$. This compares to five combinations that consist of a single input measurement and exclude $P_1$. This results in an expectation of more low scoring classifiers without $P_1$.

Figure \ref{fig_P1_stenosis} shows that in the case of LR, only 11 of the 32 classifiers trained using $P_1$ achieve an $F_1$ score of less than 0.54. This compares to all 31 LR classifiers trained without $P_1$ achieving an $F_1$ score of less than 0.54. In the case of SVM classifiers, only 1 combination of input measurements containing $P_1$ achieves an $F_1$ score of less than 0.54. This compares to 5 combinations of input measurements that do not contain $P_1$ that achieved an $F_1$ score of less than 0.54. When the threshold for comparison is increased to 0.6 it is found that 20 of the 32 SVM classifiers trained with $P_1$ exceed this threshold, compared to 14 of the 31 trained without $P_1$ exceeding this threshold. Similar analysis shows that the inclusion or exclusion of $Q_1$ produces similar patterns and behaviours in the $F_1$ scores produced. Thus measurements of pressure and flow-rate at the inlet of the system appear to be particularly informative in differentiating between healthy and unhealthy patients.

Another observation can be made by observing the highest scoring SVM classifiers in Figure \ref{fig_number_measurements}. The best performing classifiers include $P_1$ and a combination to determine the flow-split between the left and the right iliacs. For example, when three measurements are used, the best combination is $(Q_3, Q_1, P_1)$, which would enable the flow split to be known through mass conservation (note that compliance of the arteries is relatively small) in addition to $P_1$. This observations bears similarity to the classical inverse problem analysis presented in \cite{pant2014methodological}, where the authors show that in order to find the parameters of any arterial network, the inlet pressure and flow-splits to all the outlets should be known.

\subsubsection{Linear vs non-linear partitions}
 
Comparing the results achieved by LR and SVM classifiers in all previous analyses, it can be seen that SVM classifiers consistently achieve higher accuracy results than the LR classifiers. When using all the six input measurements the LR and SVM classifiers achieve sensitivities and specificities of 0.73 and 0.52; and 0.80 and 0.57, respectively.  Similarly, the $F_1$ scores for LR and SVM classifiers are 0.58 and 0.65, respectively. 
All SVM classifiers trained up to this point have mapped the input measurements provided to a higher order feature space through the use of radial basis function kernel. The fact that the accuracy of SVM classifiers are consistently higher than LR classifiers suggests that the partition between healthy and unhealthy VPs through the pressure and flow-rate measurement space is likely non-linear. To test the hypothesis that the increase in accuracy seen in SVM classifiers is due to this higher order mapping, an SVM classifier is trained and tested with a linear kernel. 
 It is found that an SVM classifier trained using all the six pressure and flow-rate measurements and a linear kernel produces an average sensitivity and specificity of 0.85 and 0.42 respectively over five folds of the VPD. This corresponds to an $F_1$ score of 0.53. The corresponding $F_1$ scores for LR and radial basis function SVM are  0.58 and 0.65, respectively. 
 The non-linear SVM outperforms the linear SVM and LR (also linear), thus demonstrating that a non-linear mapping is beneficial in discerning between healthy and diseased states.
 
 
\subsection{IVBC results}

\begin{figure}[tb]
\centering
\includegraphics[width=5.5in]{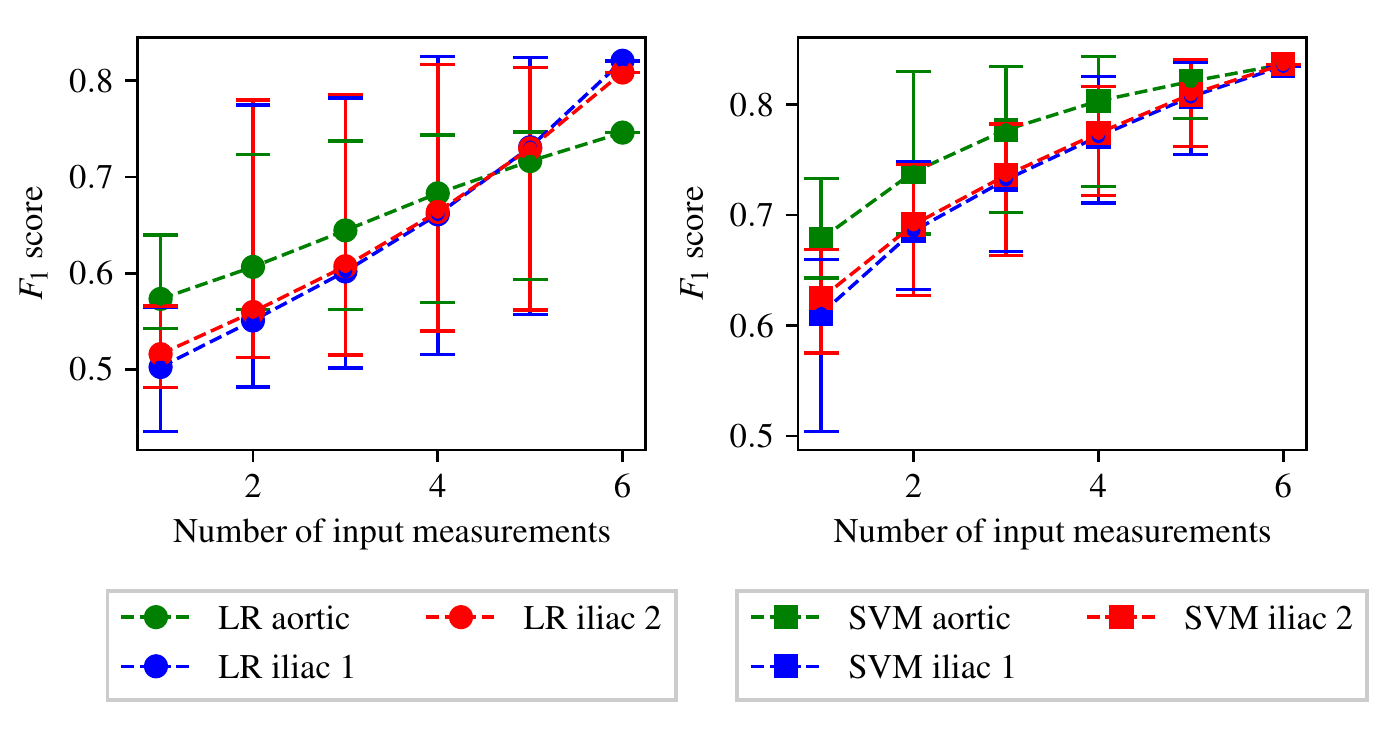}
\caption{The average, maximum, and minimum $F_1$ score achieved by all the IVBCs when providing different numbers of input measurements to detect disease in each of the three vessels. The central markers represent the average score achieved, while the error bars indicate the upper and lower limits.}
\label{fig_number_measurements_IVBC}
\end{figure}

Following an identical procedure to that employed for the ENBC combination search, three IVBC combination searches---one for disease classification in each of the three vessels---are performed using the LR and SVM methods. It is chosen to limit the IVBC combination searches to these two classification methods due to the higher computational expense, and the fact that these two methods have shown consistently higher accuracy results. The full tables of results for the IVBC combination search are presented in Appendix \ref{appendix_IVBC}. 
The average, minimum, and maximum $F_1$ score achieved when using one to six input measurements are shown in Figure \ref{fig_number_measurements_IVBC}. There is a  good agreement between the overall behaviour seen across the IVBC and ENBC (as shown in Figure \ref{fig_number_measurements}) combination searches. These similarities include:
\begin{itemize}[leftmargin=*]
\item The average and minimum $F_1$ score achieved continuously increases when increasing the number of input measurements from one to six.
\item The maximum $F_1$ score initially increases rapidly and reaches an asymptotic limit between two and four input measurements.
\item The SVM method consistently produces higher accuracy results than the LR method.
\end{itemize} 

\noindent For the SVM configurations corresponding to maximum $F_1$ scores, the sensitivities, specificities, and the combination of measurements is shown in Table \ref{table_max_min_IVBC}.
It shows that the combinations of input measurements that produce the highest $F_1$ scores in the two common iliacs are not only identical, but also symmetrical (with the same input measurements being taken from the right and left sides).  While the combinations of input measurements that produce the highest $F_1$ scores differ from the ENBC results (see Table \ref{table_max_min_svm}), a similarity between the two is that the best performing classifiers include a pressure measurement and a combination to determine the flow-split. In Section \ref{section_important_measurements} it is hypothesised that the combination of pressure at the inlet and flow-split may be particularly informative. Table \ref{table_max_min_IVBC}, however, seems to suggest that it may be the pressure within the diseased vessel and the flow-split that best captures the presence of a stenosis.

Comparing Tables \ref{table_max_min_svm} and \ref{table_max_min_IVBC} also shows that IVBCs, owing to their more granular characterisation of diseases states, lead to higher $F_1$ scores, sensitivities, and specificities, relative to the ENBCs. Neither of them are, however, good at pointing to the precise vessel that is diseased in the network. Note that even if an IVBC classifier has perfect accuracy it does not lead to knowledge of precise diseased vessel; for example, the aortic IVBC classifier only determines whether disease is in aorta, and considers both healthy and diseased iliac vessel patients together in one class (see section \ref{sec_IVBC}). When knowledge of not only the presence of disease but also the precise location is required, multiclass classifiers are necessary, and their results are presented next.


\begin{table*}
\begin{center}
\def\arraystretch{1.2}
\begin{tabular}{ |c | c | c |c  c c |}
\hline
\textbf{Number of input }  &  &  & \textbf{$F_1$} & \textbf{Sensitivity} &  \textbf{Specificity }\\
\textbf{measurements }& \textbf{Vessel} & \textbf{Combination} &\textbf{score} &  \textbf{}  &  \textbf{} \\
\hline

\multirow{3}{*}{\textbf{4}} & \textbf{Aorta}  & ($Q_3$, $Q_2$, $Q_1$, $P_1$) & 0.8437 & 0.8893 & 0.7814\\
 & \textbf{Iliac 1}  & ($Q_3$, $Q_2$, $P_3$, $P_2$) & 0.8256 & 0.8439 & 0.7996 \\
 & \textbf{Iliac 2}  & ($Q_3$, $Q_2$, $P_3$, $P_2$) & 0.8163 & 0.8303 & 0.7961 \\
\hline
\multirow{3}{*}{\textbf{5}} & \textbf{Aorta}  & ($Q_3$, $Q_2$, $Q_1$, $P_2$, $P_1$) & 0.8391 & 0.8775 & 0.7862\\
 & \textbf{Iliac 1}  & ($Q_3$, $Q_2$, $Q_1$, $P_3$, $P_2$) & 0.8387 & 0.8333 & 0.8464 \\
  & \textbf{Iliac 2}  & ($Q_3$, $Q_2$, $Q_1$, $P_3$, $P_2$) & 0.8407 & 0.8406 & 0.8409\\
\hline 
\textbf{6} & \textbf{Aorta} & ($Q_3$, $Q_2$, $Q_1$, $P_3$, $P_2$, $P_1$) & 0.8363 & 0.8734 &  0.7847\\
 & \textbf{Iliac 1} & ($Q_3$, $Q_2$, $Q_1$, $P_3$, $P_2$, $P_1$) & 0.8348 & 0.8255 &  0.8479\\
  & \textbf{Iliac 2} & ($Q_3$, $Q_2$, $Q_1$, $P_3$, $P_2$, $P_1$) & 0.8364 & 0.8276 &  0.8488\\
\hline
\end{tabular}
\caption{The combinations of input measurements that produce the maximum $F_1$ scores when providing four, five, and six input measurements to the IVBCs with the SVM method. The corresponding sensitivities and specificities are also included.}
\label{table_max_min_IVBC}
\end{center}
\end{table*}

\subsection{Multiclass analysis}
\label{section_multiclass_analysis}
Results of the muliticlass configurations are presented here. Unlike ENBC and IVBC classifier results presented above, here the goal is also to determine which vessel the disease is located in.
Due to the increased computational expense, a full combination search is not carried out for multiclass classifiers. Instead multiclass classifiers are trained and tested using the measurements of pressure and flow-rate at all the three available locations.

Initially multiclass classifiers are created using LR employing an OVA method (see section \ref{sec_ova_config}) and SVM employing an OVO method (see section \ref{sec_ovo_config}). While these initial classifiers produced high accuracy for aortic, first iliac, and second iliac disease classification, it is found that the sensitivity corresponding to the classification of VPs with `no disease' present is consistently close to 0. A multiclass classifier is, therefore, created using LR employing a CPC method, as outlined in Section \ref{sec_cpc_config}. The results of the OVA, OVO, and the CPC classifiers are shown in Table \ref{table_ova_all_comparison}. 
%
\begin{table*}
\begin{center}
\def\arraystretch{1.2}
\begin{tabular}{|c|c c |c c |c c |c c|}
\hline
 & \multicolumn{2}{c|}{\textbf{Healthy}} & \multicolumn{2}{c|}{\textbf{Aorta}} & \multicolumn{2}{c|}{\textbf{First iliac}} & \multicolumn{2}{c|}{\textbf{Second iliac}} \\
\textbf{Configuration} & \textbf{Sen.} & \textbf{Spec.} & \textbf{Sen.} & \textbf{Spec.} & \textbf{Sen.} & \textbf{Spec.}  & \textbf{Sen.} & \textbf{Spec.}\\
\hline
One-versus-all (OVA) & 0.056 & 0.986 & 0.642 & 0.851 & 0.846 & 0.729 & 0.825 & 0.725\\
One-versus-one (OVO) & 0.120 & 0.916 & 0.493 & 0.798 & 0.584 & 0.726 & 0.550 & 0.725 \\
Custom probabilistic config (CPC) & 0.496 & 0.832 & 0.581 & 0.882 & 0.745 & 0.867 & 0.722 & 0.860 \\
\hline
\end{tabular}
\caption{Multiclass accuracies of OVA, OVO, and CPC  when using pressure and flow-rate at all the three locations.}
\label{table_ova_all_comparison}
\end{center}
\end{table*}

Table \ref{table_ova_all_comparison} shows that for OVA the sensitivities and specificities for the first and second iliac are equivalent. For aorta, the specificity is relatively higher but comes at a compromise of reduced sensitivity. Finally, the healthy classification sensitivity is poor, almost close to zero. Similar behaviour is observed for OVO with almost all classification accuracies  lower when compared to OVA. Thus, OVA outperforms OVO in all cases and is thus superior for this application. When comparing OVA against CPC, the highest improvement is seen for the sensitivity of healthy classification, an increase to $\sim$50\% compared in CPC compared to $\sim$0\% in OVA. For the aorta and iliacs, a rebalancing of sensitivities and specificities is observed in relation to OVA---an increase in sensitivity is accompanied by a decrease in specificity, with their averages relatively unchanged. Overall, Table \ref{table_ova_all_comparison} shows that the CPC achieves its purpose of improving the classification accuracy for healthy (`no disease') class without significantly compromising other classification accuracies.

%

When creating CPC multiclass classifiers, preference can be given to healthy or unhealthy VPs by adjusting the decision boundary $\mathcal{B}$ in equation \eqref{eq_CPC} --- i.e. the certainty required to override the default classification that a VP has no disease present. Reducing the certainty required to change the classification a VP is assigned to, \textit{i.e.} lowering the decision boundary, creates bias towards unhealthy VPs as the CPC classifier is more willing to override the default classification that a VP is healthy. Increasing the decision boundary will require more certainty to classify a VP as diseased, giving bias toward healthy VPs, as the CPC is less willing to override the default classification that a VP is healthy.

To analyse the aforementioned affect of the decision boundary used on the classification of VPs, receiver operating characteristic (ROC) curves \cite{akobeng2007understanding} are plotted. 
ROC curves are obtained by plotting the true positive rates against the false positive rates of each  classification when different decision boundaries are applied. By recording a series of discrete true-positive/false-positive points for various decision boundaries, a curve is fitted that shows the characteristics of the accuracy of each classification across all possible decision boundaries. A complete ROC curve must start at the point representing a true positive and false positive rate of 0, \textit{i.e.} no VPs are predicted to belong to the  classification being examined, and must end at the point representing a true positive and false positive rate of 1, \textit{i.e.} all VPs are predicted to belong to the discrete classification being examined. A naive classifier, achieving an accuracy of 50\%, will produce a straight line between these two points, and so the area under the curve (AUC) is equal to 0.5. A perfect classifier ascends vertically along the the y-axis between the points (0, 0) and (0, 1), then transverses the x axis between the points (0, 1) and (1, 1). This will result in a perfect AUC score of 1. The point (0, 1) represents a perfect classifier, as  100\% of positive VPs are correctly classification, while 0\% of negative patients are incorrectly classified.
   \begin{figure}[tb]
\centering
\includegraphics[width=2.5in]{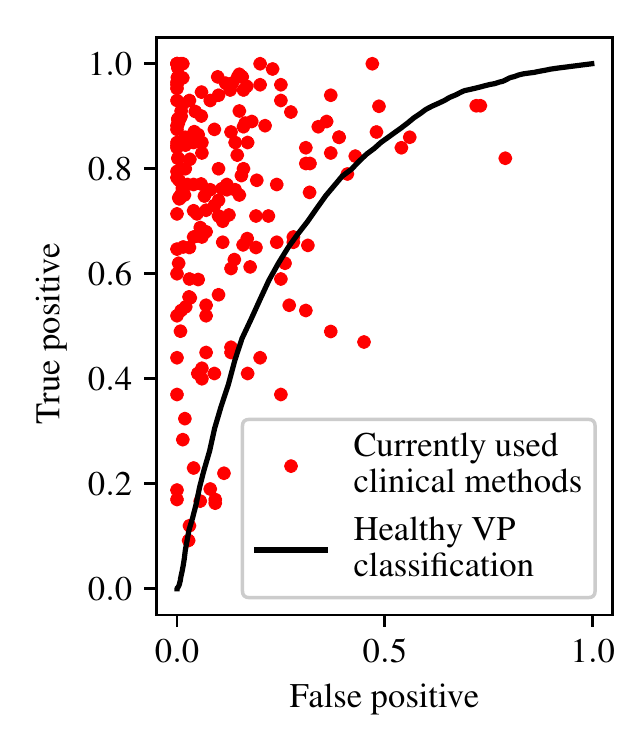}
\caption{The ROC curve of healthy VP classification within the CPC ensemble trained and tested using pressure and flow-rate measurements at all the three measurement locations. The accuracy of classification is compared to current screening methods (shown in red) \cite{akobeng2007understanding}.}
\label{fig_ROC_PAE}
\end{figure}
 Within the context of the multiclass CPC, when a decision boundary of 1 is applied, all VPs are classified as  healthy, and so the ROC curves of aortic disease, first iliac disease, and second iliac disease classification all begin at the true positive and false positive position (0, 0). When a decision boundary of 1 is applied, the true positive and false positive position of healthy classification is (1, 1), \textit{i.e.} all VPs are being assigned to the classification of no disease present. When the decision boundary is set to be 0 all VPs are classified as having disease in one of the three vessels, and so the healthy classification will reach the point (0, 0). A complete ROC curve can, therefore, be obtained for healthy classification accuracy. When the decision boundary is set to be 0, while all VPs are classified as having disease in one of the three vessels, this does not necessarily ensure that the true positive and false positive classification accuracy is equal to 1 for disease classification in each individual vessel. Complete ROC curves can, therefore, not be plotted for aortic, first iliac, and second iliac disease classification.
 %
 The ROC curve of healthy VP classification accuracy is plotted against the reported true positive and false positive rates of 193 current screening methods, recorded in \cite{maxim2014screening} and \cite{alberg2004use}, and is shown in Figure \ref{fig_ROC_PAE}.

Figure \ref{fig_ROC_PAE} shows that the ROC curve of healthy VP classification follows a  desirable profile. The AUC of the ROC curve is computed 0.75. 
An AUC of between 0.7 and 0.9 can be considered as moderate accuracy \cite{fischer2003readers}. The overall correct classification of healthy VPs by the CPC outperforms approximately 20 of the current methods.
 
The ROC curve of classifiers created in this PoC study can not be fairly compared to current screening methods, as the affects of simplifications such as only using a simple three vessel system, and limiting the number of diseased vessels to one are not understood. However, Figure  \ref{fig_ROC_PAE} provides some indication of how the results achieved in this PoC study compare to currently used screening methods. Overall, the results, despite simplifications and assumptions used in this study, are encouraging and point towards the potential of increased classification accuracies when larger networks and more sophisticated ML or deep learning algorithms are used.

\section{Conclusions}
This is the first-of-its-kind PoC study exploring ML application to detection of stenosis in arterial networks. The key conclusion is that ML methods are appropriate for detection of arterial disease, as demonstrated in the three-vessel network. The most balanced classifier, the CPC, achieves specificities larger than 80\% and sensitivities ranging from 50--75\%. The AUC under the ROC for this classifier is 0.75, which outperforms approximately 20 of the current methods used in clinical practice. This observation, motivates further exploration of more sophisticated ML and deep learning methods on virtual databases created on larger networks. This can facilitate home monitoring of disease and/or larger-scale, cost-effective, screening programmes.

Among the four ML methods considered, it is shown that LR and SVM perform significantly better than NB and RF, with the further advantage that these require little to no problem-specific optimisation. While this conclusion may be specific to the network considered, evidence shows that non-linear classification methods, such as the SVM with radial basis functions, performs better than linear classification methods for arterial disease detection. Finally, it is demonstrated that the standard methods can be modified in a custom probabilistic configuration to not only detect the presence of stenosis in the network, but also identify the diseased vessel.

This study also presents a methodological framework to both create the virtual patient database and assess that it is of adequate size for the ML applications.  The conclusion from ML classifier performance is that all measurements are not equally informative, and that similar classification accuracies can be achieved by using fewer measurements as long as the measurements are chosen judiciously. In this context, for the arterial network considered, the importance of inlet pressure, inlet flow-rate, and flow-splitat the bifurcation is highlighted.

\section{Limitations}

Several simplifications and assumptions are made during both the creation of the VPD and the training and testing of ML classifiers. These are likely to affect the classification accuracies achieved within this study. Some of these major limitations are:
\begin{itemize}[leftmargin=*]
\item The arterial network, containing only three vessels, is small. It is not clear whether this aids or hinders classification. On the one hand, due to small nature of the network, the signals are less diffused, and on the other hand specific features which may be result of uniques reflections in certain anatomical locations is not accounted for. However, the small arterial network does allow for a preliminary analysis which, with encouraging results, points towards exploration in larger networks.
\item The distribution of all arterial network parameters, excluding disease conditions, across the VPD are described using independent  distributions. These simple distributions ignore the complex inter-parameter relationships likely seen within real arterial networks. The simplification of the distribution of arterial network parameters likely results in a wider range of pressure and flow-rate profiles across the VPD, making distinction between healthy and unhealthy VPs more difficult. This may be potentially solved by first determining the probability distributions through an inverse problem approach, for example markov chain monte carlo.
\item  This  study is completed without significant consideration for clinical requirements. For example which measurements are really obtainable eaasily, and what range of stenosis severities should a ML classifier be able to detect? These questions are best explored on a larger network.
\end{itemize} 

\section*{Funding}
This work is supported by an EPSRC studentship ref. EP/N509553/1 and an EPSRC grant ref. EP/R010811/1.



\tiny
\printbibliography

\clearpage

\appendix

\footnotesize

\section{ENBC combination search results}
\label{appendix_A}

Tables \ref{table_F1_stenosis_binary}, \ref{table_healthy_stenosis_binary}, and \ref{table_unhealthy_stenosis_binary} show $F_1$ scores, sensitivities, and specificities, respectively, for the full results for ENBC combination search.
\begin{longtable}{|c|c c c c|}

\hline
& \multicolumn{4}{c|}{\textbf{Classification method}}\\
\textbf{Input combination} & \textbf{NB} & \textbf{LR} & \textbf{SVM} & \textbf{RF}\\
\hline
\endhead
\hline
\endfoot
\endlastfoot
 $Q_3$ & 0.3494 & 0.4915 & 0.5138 & 0.4789 \\
 $Q_2$ & 0.5318 & 0.4824 & 0.4989 & 0.4965 \\
 $Q_1$ & 0.3008 & 0.4932 & 0.5621 & 0.4540 \\
 $P_3$ & 0.4328 & 0.4908 & 0.5292 & 0.4926 \\
 $P_2$ & 0.4413 & 0.5060 & 0.5287 & 0.5122 \\
 $P_1$ & 0.3059 & 0.4924 & 0.5307 & 0.4705 \\
 $Q_3$, $Q_2$ & 0.4930 & 0.4878 & 0.5510 & 0.4852 \\ 
 $Q_3$, $Q_1$ & 0.3126 & 0.5136 & 0.6015 & 0.4756 \\
 $Q_3$, $P_3$ & 0.4244 & 0.4989 & 0.5710 & 0.5053 \\
 $Q_3$, $P_2$ & 0.4342 & 0.5032 & 0.5757 & 0.5109 \\  
 $Q_3$, $P_1$ & 0.3208 & 0.5077 & 0.5801 & 0.4910 \\
 $Q_2$, $Q_1$ & 0.4228 & 0.4962 & 0.5892 & 0.4916 \\  
 $Q_2$, $P_3$ & 0.4916 & 0.5057 & 0.5559 & 0.5080 \\  
 $Q_2$, $P_2$ & 0.4934 & 0.5081 & 0.5479 & 0.5046 \\  
 $Q_2$, $P_1$ & 0.3997 & 0.5054 & 0.5570 & 0.4861 \\  
 $Q_1$, $P_3$ & 0.3698 & 0.5163 & 0.6050 & 0.4956 \\
 $Q_1$, $P_2$ & 0.3806 & 0.5316 & 0.6121 & 0.5086 \\
 $Q_1$, $P_1$ & 0.3140 & 0.5190 & 0.6152 & 0.4776 \\
 $P_3$, $P_2$ & 0.4378 & 0.5052 & 0.5391 & 0.5200 \\
 $P_3$, $P_1$ & 0.3668 & 0.5267 & 0.5617 & 0.5065 \\
 $P_2$, $P_1$ & 0.3729 & 0.5397 & 0.5620 & 0.5106 \\   
 $Q_3$, $Q_2$, $Q_1$ & 0.4181 & 0.5091 & 0.6098 & 0.4901 \\
 $Q_3$, $Q_2$, $P_3$ & 0.4739 & 0.5079 & 0.5883 & 0.5080 \\
 $Q_3$, $Q_2$, $P_2$ & 0.4778 & 0.5092 & 0.5824 & 0.5090 \\
 $Q_3$, $Q_2$, $P_1$ & 0.3957 & 0.5104 & 0.5918 & 0.4945 \\  
 $Q_3$, $Q_1$, $P_3$ & 0.3728 & 0.5292 & 0.6240 & 0.4957 \\
 $Q_3$, $Q_1$, $P_2$ & 0.3840 & 0.5292 & 0.6279 & 0.5138 \\
 $Q_3$, $Q_1$, $P_1$ & 0.3205 & 0.5290 & 0.6356 & 0.4909 \\
 $Q_3$, $P_3$, $P_2$ & 0.4360 & 0.5041 & 0.5769 & 0.5085 \\
 $Q_3$, $P_3$, $P_1$ & 0.3702 & 0.5420 & 0.5983 & 0.5049 \\
 $Q_3$, $P_2$, $P_1$ & 0.3770 & 0.5435 & 0.5985 & 0.5160\\
 $Q_2$, $Q_1$, $P_3$ & 0.4444 & 0.5223 & 0.6117 & 0.5036\\
 $Q_2$, $Q_1$, $P_2$ & 0.4437 & 0.5348 & 0.6105 & 0.5013\\
 $Q_2$, $Q_1$, $P_1$ & 0.3780 & 0.5262 & 0.6182 & 0.4901\\
 $Q_2$, $P_3$, $P_2$ & 0.4741 & 0.5090 & 0.5629 & 0.5179\\ 
 $Q_2$, $P_3$, $P_1$ & 0.4119 & 0.5378 & 0.5769 & 0.5103\\
 $Q_2$, $P_2$, $P_1$ & 0.4165 & 0.5478 & 0.5761 & 0.5153\\
 $Q_1$, $P_3$, $P_2$ & 0.4121 & 0.5344 & 0.6143 & 0.5196\\
 $Q_1$, $P_3$, $P_1$ & 0.3433 & 0.5470 & 0.6221 & 0.4948\\
 $Q_1$, $P_2$, $P_1$ & 0.3507 & 0.5549 & 0.6228 & 0.5146\\
 $P_3$, $P_2$, $P_1$ & 0.3938 & 0.5518 & 0.5740 & 0.5239\\
 $Q_3$, $Q_2$, $Q_1$, $P_3$ & 0.4376 & 0.5292 & 0.6280 & 0.5023\\ 
 $Q_3$, $Q_2$, $Q_1$, $P_2$ & 0.4395 & 0.5384 & 0.6273 & 0.5060\\   
 $Q_3$, $Q_2$, $Q_1$, $P_1$ & 0.3797 & 0.5350 & 0.6368 & 0.4947\\
 $Q_3$, $Q_2$, $P_3$, $P_2$ & 0.4696 & 0.5098 & 0.5929 & 0.5183\\   
 $Q_3$, $Q_2$, $P_3$, $P_1$ & 0.4105 & 0.5466 & 0.6100 & 0.5119\\   
 $Q_3$, $Q_2$, $P_2$, $P_1$ & 0.4144 & 0.5482 & 0.6078 & 0.5175\\   
 $Q_3$, $Q_1$, $P_3$, $P_2$ & 0.4104 & 0.5364 & 0.6240 & 0.5079\\   
 $Q_3$, $Q_1$, $P_3$, $P_1$ & 0.3516 & 0.5588 & 0.6429 & 0.5043\\   
 $Q_3$, $Q_1$, $P_2$, $P_1$ & 0.3540 & 0.5587 & 0.6407 & 0.5103\\
 $Q_3$, $P_3$, $P_2$, $P_1$ & 0.3956 & 0.5596 & 0.6025 & 0.5163\\  
 $Q_2$, $Q_1$, $P_3$, $P_2$ & 0.4511 & 0.5387 & 0.6153 & 0.5093\\ 
 $Q_2$, $Q_1$, $P_3$, $P_1$ & 0.3956 & 0.5538 & 0.6268 & 0.5053\\
 $Q_2$, $Q_1$, $P_2$, $P_1$ & 0.3982 & 0.5676 & 0.6274 & 0.5170\\
 $Q_2$, $P_3$, $P_2$, $P_1$ & 0.4238 & 0.5597 & 0.5836 & 0.5229\\
 $Q_1$, $P_3$, $P_2$, $P_1$ & 0.3773 & 0.5698 & 0.6277 & 0.5180\\
 $Q_3$, $Q_2$, $Q_1$, $P_3$, $P_2$ & 0.4455 & 0.5397 & 0.6275 & 0.5161\\
 $Q_3$, $Q_2$, $Q_1$, $P_3$, $P_1$ & 0.3955 & 0.5670 & 0.6469 & 0.5129\\
 $Q_3$, $Q_2$, $Q_1$, $P_2$, $P_1$ & 0.4000 & 0.5686 & 0.6432 & 0.5139\\ 
 $Q_3$, $Q_2$, $P_3$, $P_2$, $P_1$ & 0.4251 & 0.5595 & 0.6140 & 0.5127\\ 
 $Q_3$, $Q_1$, $P_3$, $P_2$, $P_1$ & 0.3791 & 0.5695 & 0.6434 & 0.5191\\ 
 $Q_2$, $Q_1$, $P_3$, $P_2$, $P_1$ & 0.4100 & 0.5718 & 0.6306 & 0.5188\\
 $Q_3$, $Q_2$, $Q_1$, $P_3$, $P_2$, $P_1$ & 0.4139 & 0.5815 & 0.6454 & 0.5246\\  
\hline 
  \caption{The $F_1$ scores achieved across the ENBC combination search by each of the four classification methods.}
 \label{table_F1_stenosis_binary}
\end{longtable}

\clearpage

\begin{longtable}{|c|c c c c|}

\hline
& \multicolumn{4}{c|}{\textbf{Classification method}}\\
\textbf{Input combination} & \textbf{NB} & \textbf{LR} & \textbf{SVM} & \textbf{RF}\\
\hline
\endhead
\hline
\endfoot
\endlastfoot
 $Q_3$ & 0.7431 & 0.5516 & 0.6868 & 0.5961\\
 $Q_2$ & 0.4624 & 0.5896 & 0.6932 & 0.5669\\
 $Q_1$ & 0.8321 & 0.5348 & 0.7154 & 0.5956\\
 $P_3$ & 0.6755 & 0.5833 & 0.7289 & 0.5654\\
 $P_2$ & 0.6732 & 0.6038 & 0.7445 & 0.5681\\
 $P_1$ & 0.8094 & 0.6309 & 0.7634 & 0.6168\\
 $Q_3$, $Q_2$ & 0.5447 & 0.5686 & 0.7186 & 0.59611\\
 $Q_3$, $Q_1$ & 0.8127 & 0.5355 & 0.7220 & 0.6413\\
 $Q_3$, $P_3$ & 0.6817 & 0.5738 & 0.7144 & 0.5741\\
 $Q_3$, $P_2$ & 0.6710 & 0.5928 & 0.7183 & 0.5704\\
 $Q_3$, $P_1$ & 0.7803 & 0.6239 & 0.7603 & 0.6121\\
 $Q_2$, $Q_1$ & 0.6912 & 0.5684 & 0.7387 & 0.6150\\
 $Q_2$, $P_3$ & 0.5907 & 0.5840 & 0.7311 & 0.5791\\
 $Q_2$, $P_2$ & 0.5941 & 0.5879 & 0.7466 & 0.5812\\
 $Q_2$, $P_1$ & 0.7303 & 0.6213 & 0.8000 & 0.6344\\
 $Q_1$, $P_3$ & 0.7664 & 0.5754 & 0.7532 & 0.5930\\
 $Q_1$, $P_2$ & 0.7657 & 0.5946 & 0.7595 & 0.5926\\
 $Q_1$, $P_1$ & 0.8299 & 0.6406 & 0.7934 & 0.6283\\
 $P_3$, $P_2$ & 0.6731 & 0.5984 & 0.7402 & 0.5729\\
 $P_3$, $P_1$ & 0.7607 & 0.7386 & 0.8009 & 0.6027\\
 $P_2$, $P_1$ & 0.7631 & 0.7349 & 0.8067 & 0.6047\\
 $Q_3$, $Q_2$, $Q_1$ & 0.6952 & 0.5706 & 0.7693 & 0.6200\\
 $Q_3$, $Q_2$, $P_3$ & 0.6100 & 0.5784 & 0.7379 & 0.5835\\
 $Q_3$, $Q_2$, $P_2$ & 0.6075 & 0.5798 & 0.7378 & 0.5880\\
 $Q_3$, $Q_2$, $P_1$ & 0.7167 & 0.6201 & 0.7854 & 0.6255\\
 $Q_3$, $Q_1$, $P_3$ & 0.7560 & 0.5708 & 0.7516 & 0.6052\\
 $Q_3$, $Q_1$, $P_2$ & 0.7507 & 0.6032 & 0.7607 & 0.6034\\
 $Q_3$, $Q_1$, $P_1$ & 0.8129 & 0.6330 & 0.7857 & 0.6217\\
 $Q_3$, $P_3$, $P_2$ & 0.6699 & 0.5949 & 0.7297 & 0.5813\\
 $Q_3$, $P_3$, $P_1$ & 0.7507 & 0.7209 & 0.7966 & 0.6173\\
 $Q_3$, $P_2$, $P_1$ & 0.7485 & 0.7198 & 0.7910 & 0.6094\\
 $Q_2$, $Q_1$, $P_3$ & 0.6925 & 0.5963 & 0.7723 & 0.5982\\
 $Q_2$, $Q_1$, $P_2$ & 0.6950 & 0.5896 & 0.7773 & 0.5999\\
 $Q_2$, $Q_1$, $P_1$ & 0.7711 & 0.6376 & 0.8059 & 0.6420\\
 $Q_2$, $P_3$, $P_2$ & 0.6308 & 0.5890 & 0.7418 & 0.5811\\
 $Q_2$, $P_3$, $P_1$ & 0.7187 & 0.7151 & 0.7992 & 0.6095\\
 $Q_2$, $P_2$, $P_1$ & 0.7181 & 0.7290 & 0.8165 & 0.6140\\
 $Q_1$, $P_3$, $P_2$ & 0.7330 & 0.5935 & 0.7623 & 0.5968\\
 $Q_1$, $P_3$, $P_1$ & 0.7951 & 0.7096 & 0.7934 & 0.6220\\
 $Q_1$, $P_2$, $P_1$ & 0.7926 & 0.7060 & 0.8023 & 0.6264\\
 $P_3$, $P_2$, $P_1$ & 0.7391 & 0.7388 & 0.8016 & 0.5958\\
 $Q_3$, $Q_2$, $Q_1$, $P_3$ & 0.6903 & 0.6116 & 0.7872 & 0.6062\\
 $Q_3$, $Q_2$, $Q_1$, $P_2$ & 0.6872 & 0.6169 & 0.7861 & 0.5911\\
 $_3$, $Q_2$, $Q_1$, $P_1$ & 0.7593 & 0.6470 & 0.8098 & 0.6410\\
 $Q_3$, $Q_2$, $P_3$, $P_2$ & 0.6325 & 0.5846 & 0.7370 & 0.5868\\
 $Q_3$, $Q_2$, $P_3$, $P_1$ & 0.7089 & 0.7115 & 0.7963 & 0.6096\\
 $Q_3$, $Q_2$, $P_2$, $P_1$ & 0.7081 & 0.7219 & 0.7970 & 0.6134\\
 $Q_3$, $Q_1$, $P_3$, $P_2$ & 0.7266 & 0.6026 & 0.7680 & 0.6088\\
 $Q_3$, $Q_1$, $P_3$, $P_1$ & 0.7760 & 0.6973 & 0.7994 & 0.6221\\
 $Q_3$, $Q_1$, $P_2$, $P_1$ & 0.7754 & 0.6911 & 0.7965 & 0.6211\\
 $Q_3$, $P_3$, $P_2$, $P_1$ & 0.7314 & 0.7321 & 0.7962 & 0.6056\\
 $Q_2$, $Q_1$, $P_3$, $P_2$ & 0.6908 & 0.6039 & 0.7773 & 0.5978\\
 $Q_2$, $Q_1$, $P_3$, $P_1$ & 0.7517 & 0.6999 & 0.8006 & 0.6297\\
 $Q_2$, $Q_1$, $P_2$, $P_1$ & 0.7514 & 0.7156 & 0.8152 & 0.6207\\
 $Q_2$, $P_3$, $P_2$, $P_1$ & 0.7081 & 0.7386 & 0.8059 & 0.6005\\
 $Q_1$, $P_3$, $P_2$, $P_1$ & 0.7682 & 0.7115 & 0.7982 & 0.6213\\
 $Q_3$, $Q_2$, $Q_1$, $P_3$, $P_2$ & 0.6872 & 0.6186 & 0.7858 & 0.6013\\
 $Q_3$, $Q_2$, $Q_1$, $P_3$, $P_1$ & 0.7402 & 0.7121 & 0.8115 & 0.6127\\
 $Q_3$, $Q_2$, $Q_1$, $P_2$, $P_1$ & 0.7394 & 0.7175 & 0.8103 & 0.6386\\
 $Q_3$, $Q_2$, $P_3$, $P_2$, $P_1$ & 0.7022 & 0.7310 & 0.7947 & 0.6029\\
 $Q_3$, $Q_1$, $P_3$, $P_2$, $P_1$ & 0.7587 & 0.7016 & 0.7968 & 0.6182\\
 $Q_2$, $Q_1$, $P_3$, $P_2$, $P_1$ & 0.7390 & 0.7234 & 0.8022 & 0.6149\\
 $Q_3$, $Q_2$, $Q_1$, $P_3$, $P_2$, $P_1$ & 0.7322 & 0.7267 & 0.8050 & 0.6106\\

\hline 
  \caption{The sensitivities achieved across the ENBC combination search by each of the four classification methods.}
 \label{table_healthy_stenosis_binary}
\end{longtable}

\clearpage

\begin{longtable}{|c|c  c c c|}

\hline
& \multicolumn{4}{c|}{\textbf{Classification method}}\\
\textbf{Input combination} & \textbf{NB} & \textbf{LR} & \textbf{SVM} & \textbf{RF}\\
\hline
\endhead
\hline
\endfoot
\endlastfoot
 $Q_3$ & 0.2660 & 0.4720 & 0.4540 & 0.4421\\
 $Q_2$ & 0.5570 & 0.4484 & 0.4344 & 0.4733\\
 $Q_1$ & 0.2067 & 0.4797 & 0.5022 & 0.4125\\
 $P_3$ & 0.3657 & 0.4608 & 0.4574 & 0.4689\\
 $P_2$ & 0.3757 & 0.4730 & 0.4512 & 0.4930\\
 $P_1$ & 0.2150 & 0.4472 & 0.4467 & 0.4255\\
 $Q_3$, $Q_2$ & 0.4762 & 0.4618 & 0.4873 & 0.4497\\
 $Q_3$, $Q_1$ & 0.2199 & 0.5061 & 0.5497 & 0.4239\\
 $Q_3$, $P_3$ & 0.3550 & 0.4741 & 0.5137 & 0.4821\\
 $Q_3$, $P_2$ & 0.3685 & 0.4732 & 0.5181 & 0.4906\\
 $Q_3$, $P_1$ & 0.2331 & 0.4682 & 0.5065 & 0.4517\\
 $Q_2$, $Q_1$ & 0.3508 & 0.4724 & 0.5268 & 0.4515\\
 $Q_2$, $P_3$ & 0.4594 & 0.4793 & 0.4885 & 0.4838\\
 $Q_2$, $P_2$ & 0.4604 & 0.4810 & 0.4729 & 0.4788\\
 $Q_2$, $P_1$ & 0.3171 & 0.4663 & 0.4633 & 0.4385\\
 $Q_1$, $P_3$ & 0.2799 & 0.4958 & 0.5407 & 0.4635\\
 $Q_1$, $P_2$ & 0.2900 & 0.5089 & 0.5471 & 0.4800\\
 $Q_1$, $P_1$ & 0.2179 & 0.4764 & 0.5361 & 0.4304\\
 $P_3$, $P_2$ & 0.3718 & 0.4738 & 0.4649 & 0.5015\\
 $P_3$, $P_1$ & 0.2784 & 0.4510 & 0.4683 & 0.4740\\
 $P_2$, $P_1$ & 0.2834 & 0.4676 & 0.4664 & 0.4784\\
 $Q_3$, $Q_2$, $Q_1$ & 0.3448 & 0.4882 & 0.5399 & 0.4480\\
 $Q_3$, $Q_2$, $P_3$ & 0.4317 & 0.4840 & 0.5260 & 0.4823\\
 $Q_3$, $Q_2$, $P_2$ & 0.4372 & 0.4852 & 0.5186 & 0.4820\\
 $Q_3$, $Q_2$, $P_1$ & 0.3166 & 0.4729 & 0.5104 & 0.4515\\
 $Q_3$, $Q_1$, $P_3$ & 0.2850 & 0.5143 & 0.5662 & 0.4597\\
 $Q_3$, $Q_1$, $P_2$ & 0.2969 & 0.5027 & 0.5672 & 0.4829\\
 $Q_3$, $Q_1$, $P_1$ & 0.2265 & 0.4917 & 0.5658 & 0.4484\\
 $Q_3$, $P_3$, $P_2$ & 0.3707 & 0.4736 & 0.5150 & 0.4838\\
 $Q_3$, $P_3$, $P_1$ & 0.2837 & 0.4756 & 0.5137 & 0.4670\\ 
 $Q_3$, $P_2$, $P_1$ & 0.2908 & 0.4778 & 0.5164 & 0.4835\\
 $Q_2$, $Q_1$, $P_3$ & 0.3734 & 0.4962 & 0.5410 & 0.4718\\
 $Q_2$, $Q_1$, $P_2$ & 0.3721 & 0.5149 & 0.5373 & 0.4684\\
 $Q_2$, $Q_1$, $P_1$ & 0.2863 & 0.4865 & 0.5343 & 0.4408\\
 $Q_2$, $P_3$, $P_2$ & 0.4254 & 0.4817 & 0.4929 & 0.4959\\
 $Q_2$, $P_3$, $P_1$ & 0.3324 & 0.4726 & 0.4869 & 0.4764\\
 $Q_2$, $P_2$, $P_1$ & 0.3371 & 0.4795 & 0.4789 & 0.4811\\
 $Q_1$, $P_3$, $P_2$ & 0.3288 & 0.5129 & 0.5487 & 0.4926\\
 $Q_1$, $P_3$, $P_1$ & 0.2497 & 0.4859 & 0.5448 & 0.4530\\
 $Q_1$, $P_2$, $P_1$ & 0.2568 & 0.4970 & 0.5417 & 0.4759\\
 $P_3$, $P_2$, $P_1$ & 0.3092 & 0.4806 & 0.4825 & 0.4985\\
 $Q_3$, $Q_2$, $Q_1$, $P_3$ & 0.3668 & 0.4996 & 0.5551 & 0.4675\\
 $Q_3$, $Q_2$, $Q_1$, $P_2$ & 0.3697 & 0.5095 & 0.5548 & 0.4773\\
 $Q_3$, $Q_2$, $Q_1$, $P_1$ & 0.2908 & 0.4942 & 0.5560 & 0.4466\\
 $Q_3$, $Q_2$, $P_3$, $P_2$ & 0.4196 & 0.4843 & 0.5322 & 0.4944\\
 $Q_3$, $Q_2$, $P_3$, $P_1$ & 0.3335 & 0.4847 & 0.5282 & 0.4784\\
 $Q_3$, $Q_2$, $P_2$, $P_1$ & 0.3376 & 0.4827 & 0.5252 & 0.4841\\
 $Q_3$, $Q_1$, $P_3$, $P_2$ & 0.3287 & 0.5122 & 0.5587 & 0.4736\\
 $Q_3$, $Q_1$, $P_3$, $P_1$ & 0.2610 & 0.5051 & 0.5688 & 0.4646\\
 $Q_3$, $Q_1$, $P_2$, $P_1$ & 0.2634 & 0.5074 & 0.5673 & 0.4724\\
 $Q_3$, $P_3$, $P_2$, $P_1$ & 0.3127 & 0.4927 & 0.5190 & 0.4853\\
 $Q_2$, $Q_1$, $P_3$, $P_2$ & 0.3813 & 0.5147 & 0.5434 & 0.4791\\
 $Q_2$, $Q_1$, $P_3$, $P_1$ & 0.3078 & 0.4979 & 0.5475 & 0.4633\\
 $Q_2$, $Q_1$, $P_2$, $P_1$ & 0.3104 & 0.5090 & 0.5416 & 0.4809\\
 $Q_2$, $P_3$, $P_2$, $P_1$ & 0.3473 & 0.4903 & 0.4920 & 0.4954\\
 $Q_1$, $P_3$, $P_2$, $P_1$ & 0.2865 & 0.5134 & 0.5497 & 0.4819\\
 $Q_3$, $Q_2$, $Q_1$, $P_3$, $P_2$ & 0.3762 & 0.5106 & 0.5552 & 0.4865\\
 $Q_3$, $Q_2$, $Q_1$, $P_3$, $P_1$ & 0.3105 & 0.5096 & 0.5683 & 0.4785\\
 $Q_3$, $Q_2$, $Q_1$, $P_2$, $P_1$ & 0.3151 & 0.5095 & 0.5640 & 0.4708\\
 $Q_3$, $Q_2$, $P_3$, $P_2$, $P_1$ & 0.3504 & 0.4930 & 0.5340 & 0.4816\\
 $Q_3$, $Q_1$, $P_3$, $P_2$, $P_1$ & 0.2902 & 0.5170 & 0.5707 & 0.4844\\
 $Q_2$, $Q_1$, $P_3$, $P_2$, $P_1$ & 0.3251 & 0.5112 & 0.5516 & 0.4852\\
 $Q_3$, $Q_2$, $Q_1$, $P_3$, $P_2$, $P_1$ &  0.3308 & 0.5221 & 0.5694 & 0.4941\\

\hline 
  \caption{The specificities achieved across the ENBC combination search by each of the four classification methods.}
 \label{table_unhealthy_stenosis_binary}
\end{longtable}

\newpage

\section{IVBC combination search results}
Tables \ref{table_F1_IVBC}, \ref{table_sens_IVBC}, and \ref{table_spec_IVBC} show $F_1$ scores, sensitivities, and specificities, respectively, for the full results for ENBC combination search.
\label{appendix_IVBC}

\begin{longtable}{|c|c c c | c c c|}

\hline
& \multicolumn{3}{c|}{\textbf{LR}} & \multicolumn{3}{c|}{\textbf{SVM}}\\
\textbf{Input combination} & \textbf{Aortic} & \textbf{Iliac 1} & \textbf{Iliac 2} & \textbf{Aortic} & \textbf{Iliac 1} & \textbf{Iliac 2} \\
\hline
\endhead
\hline
\endfoot
\endlastfoot
 $Q_3$ & 0.5588 & 0.4356 & 0.4974 & 0.6431 & 0.5043 & 0.6056 \\
 $Q_2$ & 0.5661 & 0.4953 & 0.4810 & 0.6515 & 0.6057 & 0.5750 \\
 $Q_1$ & 0.5423 & 0.4895 & 0.5103 & 0.7010 & 0.6452 & 0.6686 \\
 $P_3$ & 0.5664 & 0.5226 & 0.5659 & 0.6700 & 0.6354 & 0.6647 \\
 $P_2$ & 0.5666 & 0.5650 & 0.5233 & 0.6716 & 0.6596 & 0.6309 \\
 $P_1$ & 0.6395 & 0.5065 & 0.5171 & 0.7332 & 0.6143 & 0.6035 \\
 $Q_3$, $Q_2$ & 0.5622 & 0.4891 & 0.5144 & 0.6909 & 0.6326 & 0.6654 \\
 $Q_3$, $Q_1$ & 0.5626 & 0.4816 & 0.5266 & 0.7323 & 0.6715 & 0.7166 \\
 $Q_3$, $P_3$ & 0.5654 & 0.5210 & 0.5631 & 0.6939 & 0.6519 & 0.6869 \\
 $Q_3$, $P_2$ & 0.5701 & 0.5759 & 0.5401 & 0.6941 & 0.7083 & 0.6894 \\
 $Q_3$, $P_1$ & 0.6391 & 0.5081 & 0.5283 & 0.7717 & 0.6444 & 0.6632 \\
 $Q_2$, $Q_1$ & 0.5629 & 0.5033 & 0.5168 & 0.7339 & 0.6878 & 0.6844 \\
 $Q_2$, $P_3$ & 0.5638 & 0.5273 & 0.5806 & 0.6981 & 0.6782 & 0.7108 \\
 $Q_2$, $P_2$ & 0.5622 & 0.5652 & 0.5205 & 0.6965 & 0.6850 & 0.6528 \\
 $Q_2$, $P_1$ & 0.6405 & 0.5208 & 0.5199 & 0.7733 & 0.6584 & 0.6534 \\
 $Q_1$, $P_3$ & 0.5832 & 0.5226 & 0.5834 & 0.7586 & 0.7118 & 0.7456 \\
 $Q_1$, $P_2$ & 0.5893 & 0.5865 & 0.5320 & 0.7590 & 0.7448 & 0.7098 \\
 $Q_1$, $P_1$ & 0.6843 & 0.5040 & 0.5125 & 0.8301 & 0.6996 & 0.7059 \\
 $P_3$, $P_2$ & 0.5658 & 0.7746 & 0.7800 & 0.6829 & 0.7478 & 0.7437 \\
 $P_3$, $P_1$ & 0.7233 & 0.5425 & 0.6456 & 0.7853 & 0.6477 & 0.7149 \\
 $P_2$, $P_1$ & 0.7235 & 0.6392 & 0.5303 & 0.7854 & 0.7156 & 0.6270 \\
 $Q_3$, $Q_2$, $Q_1$ & 0.5628 & 0.5014 & 0.5374 & 0.7572 & 0.7192 & 0.7422 \\
 $Q_3$, $Q_2$, $P_3$ & 0.5651 & 0.5369 & 0.5783 & 0.7221 & 0.7069 & 0.7328 \\
 $Q_3$, $Q_2$, $P_2$ & 0.5675 & 0.5754 & 0.5498 & 0.7210 & 0.7355 & 0.7175 \\
 $Q_3$, $Q_2$, $P_1$ & 0.6417 & 0.5144 & 0.5369 & 0.7935 & 0.6844 & 0.7020 \\
 $Q_3$, $Q_1$, $P_3$ & 0.5806 & 0.5271 & 0.5794 & 0.7693 & 0.7267 & 0.7687 \\
 $Q_3$, $Q_1$, $P_2$ & 0.5949 & 0.6066 & 0.5367 & 0.7739 & 0.7763 & 0.7475 \\
 $Q_3$, $Q_1$, $P_1$ & 0.6844 & 0.5028 & 0.5327 & 0.8346 & 0.7149 & 0.7409 \\
 $Q_3$, $P_3$, $P_2$ & 0.5745 & 0.7821 & 0.7680 & 0.7024 & 0.7825 & 0.7728 \\
 $Q_3$, $P_3$, $P_1$ & 0.7300 & 0.5465 & 0.6477 & 0.7980 & 0.6670 & 0.7329 \\
 $Q_3$, $P_2$, $P_1$ & 0.7201 & 0.6455 & 0.5400 & 0.7940 & 0.7368 & 0.6892 \\
 $Q_2$, $Q_1$, $P_3$ & 0.5881 & 0.5240 & 0.6055 & 0.7745 & 0.7331 & 0.7824 \\
 $Q_2$, $Q_1$, $P_2$ & 0.5864 & 0.5815 & 0.5354 & 0.7701 & 0.7546 & 0.7239 \\
 $Q_2$, $Q_1$, $P_1$ & 0.6901 & 0.5169 & 0.5151 & 0.8329 & 0.7246 & 0.7182 \\
 $Q_2$, $P_3$, $P_2$ & 0.5620 & 0.7650 & 0.7857 & 0.7076 & 0.7680 & 0.7721 \\
 $Q_2$, $P_3$, $P_1$ & 0.7220 & 0.5492 & 0.6507 & 0.7953 & 0.6869 & 0.7366 \\
 $Q_2$, $P_2$, $P_1$ & 0.7373 & 0.6330 & 0.5290 & 0.7960 & 0.7272 & 0.6632 \\
 $Q_1$, $P_3$, $P_2$ & 0.5926 & 0.7646 & 0.7647 & 0.7631 & 0.7775 & 0.7717 \\
 $Q_1$, $P_3$, $P_1$ & 0.7291 & 0.5439 & 0.6408 & 0.8259 & 0.7084 & 0.7606 \\
 $Q_1$, $P_2$, $P_1$ & 0.7329 & 0.6481 & 0.5392 & 0.8249 & 0.7614 & 0.7052 \\
 $P_3$, $P_2$, $P_1$ & 0.7265 & 0.7698 & 0.7728 & 0.7869 & 0.7499 & 0.7414 \\
 $Q_3$, $Q_2$, $Q_1$, $P_3$ & 0.5866 & 0.5330 & 0.6016 & 0.7857 & 0.7489 & 0.8039 \\
 $Q_3$, $Q_2$, $Q_1$, $P_2$ & 0.5947 & 0.6009 & 0.5501 & 0.7858 & 0.7918 & 0.7641 \\
 $Q_3$, $Q_2$, $Q_1$, $P_1$ & 0.6898 & 0.5156 & 0.5412 & 0.8437 & 0.7388 & 0.7538 \\
 $Q_3$, $Q_2$, $P_3$, $P_2$ & 0.5693 & 0.8255 & 0.8167 & 0.7259 & 0.8256 & 0.8163 \\
 $Q_3$, $Q_2$, $P_3$, $P_1$ & 0.7288 & 0.5538 & 0.6481 & 0.8053 & 0.7107 & 0.7550 \\
 $Q_3$, $Q_2$, $P_2$, $P_1$ & 0.7358 & 0.6402 & 0.5517 & 0.8049 & 0.7557 & 0.7175 \\
 $Q_3$, $Q_1$, $P_3$, $P_2$ & 0.5975 & 0.7783 & 0.7686 & 0.7769 & 0.8188 & 0.8115 \\
 $Q_3$, $Q_1$, $P_3$, $P_1$ & 0.7358 & 0.5490 & 0.6459 & 0.8322 & 0.7266 & 0.7857 \\
 $Q_3$, $Q_1$, $P_2$, $P_1$ & 0.7352 & 0.6461 & 0.5562 & 0.8309 & 0.7851 & 0.7465 \\
 $Q_3$, $P_3$, $P_2$, $P_1$ & 0.7309 & 0.7762 & 0.7693 & 0.7967 & 0.7880 & 0.7910 \\
 $Q_2$, $Q_1$, $P_3$, $P_2$ & 0.5932 & 0.7752 & 0.7818 & 0.7789 & 0.8033 & 0.8144 \\
 $Q_2$, $Q_1$, $P_3$, $P_1$ & 0.7325 & 0.5551 & 0.6498 & 0.8310 & 0.7359 & 0.7857 \\
 $Q_2$, $Q_1$, $P_2$, $P_1$ & 0.7438 & 0.6422 & 0.5397 & 0.8322 & 0.7717 & 0.7197 \\
 $Q_2$, $P_3$, $P_2$, $P_1$ & 0.7353 & 0.7655 & 0.7771 & 0.7968 & 0.7898 & 0.7784 \\
 $Q_1$, $P_3$, $P_2$, $P_1$ & 0.7358 & 0.7606 & 0.7583 & 0.8213 & 0.7759 & 0.7707 \\
 $Q_3$, $Q_2$, $Q_1$, $P_3$, $P_2$ & 0.5932 & 0.8147 & 0.8069 & 0.7875 & 0.8387 & 0.8407 \\
 $Q_3$, $Q_2$, $Q_1$, $P_3$, $P_1$ & 0.7412 & 0.5568 & 0.6498 & 0.8387 & 0.7550 & 0.8028 \\
 $Q_3$, $Q_2$, $Q_1$, $P_2$, $P_1$ & 0.7466 & 0.6401 & 0.5616 & 0.8391 & 0.7961 & 0.7622 \\
 $Q_3$, $Q_2$, $P_3$, $P_2$, $P_1$ & 0.7352 & 0.8241 & 0.8139 & 0.8051 & 0.8219 & 0.8180 \\
 $Q_3$, $Q_1$, $P_3$, $P_2$, $P_1$ & 0.7391 & 0.7764 & 0.7728 & 0.8283 & 0.8208 & 0.8190 \\
 $Q_2$, $Q_1$, $P_3$, $P_2$, $P_1$ & 0.7440 & 0.7738 & 0.7733 & 0.8269 & 0.8096 & 0.8125 \\
 $Q_3$, $Q_2$, $Q_1$, $P_3$, $P_2$, $P_1$ & 0.7461 & 0.8208 & 0.8086 & 0.8363 & 0.8348 & 0.8364 \\
 
\hline 
  \caption{The $F_1$ scores achieved across the IVBC combination searches by the LR and SVM classification methods.}
 \label{table_F1_IVBC}
\end{longtable}

\begin{longtable}{|c|c c c | c c c|}

\hline
& \multicolumn{3}{c|}{\textbf{LR}} & \multicolumn{3}{c|}{\textbf{SVM}}\\
\textbf{Input combination} & \textbf{Aortic} & \textbf{Iliac 1} & \textbf{Iliac 2} & \textbf{Aortic} & \textbf{Iliac 1} & \textbf{Iliac 2} \\
\hline
\endhead
\hline
\endfoot
\endlastfoot
 $Q_3$ & 0.5937 & 0.3853 & 0.4850 & 0.6992 & 0.4245 & 0.5942 \\
 $Q_2$ & 0.6166 & 0.4791 & 0.4844 & 0.7289 & 0.6135 & 0.5500 \\
 $Q_1$ & 0.5691 & 0.4990 & 0.5216 & 0.7338 & 0.5971 & 0.6517 \\
 $P_3$ & 0.6065 & 0.5310 & 0.5868 & 0.7239 & 0.6156 & 0.6873 \\
 $P_2$ & 0.6096 & 0.5847 & 0.5390 & 0.7212 & 0.6410 & 0.6332 \\
 $P_1$ & 0.6932 & 0.5064 & 0.5276 & 0.7930 & 0.5808 & 0.5705 \\
 $Q_3$, $Q_2$ & 0.6020 & 0.4694 & 0.5210 &  0.7591 & 0.5777 & 0.6411 \\
 $Q_3$, $Q_1$ & 0.5861 & 0.4751 & 0.5303 &  0.7549 & 0.6045 & 0.7036 \\
 $Q_3$, $P_3$ & 0.6055 & 0.5236 & 0.5784 &  0.7258 & 0.6110 & 0.6891 \\
 $Q_3$, $P_2$ & 0.6117 & 0.6006 & 0.5435 &  0.7248 & 0.7045 & 0.6733 \\
 $Q_3$, $P_1$ & 0.6958 & 0.5048 & 0.5442 &  0.8258 & 0.5967 & 0.6457 \\
 $Q_2$, $Q_1$ & 0.5957 & 0.4969 & 0.5320 &  0.7694 & 0.6324 & 0.6541 \\
 $Q_2$, $P_3$ & 0.6086 & 0.5149 & 0.6119 &  0.7536 & 0.6376 & 0.7327 \\
 $Q_2$, $P_2$ & 0.6030 & 0.5772 & 0.5377 &  0.7538 & 0.6744 & 0.6407 \\
 $Q_2$, $P_1$ & 0.6978 & 0.5183 & 0.5374 &  0.8416 & 0.6327 & 0.6285 \\
 $Q_1$, $P_3$ & 0.6262 & 0.5325 & 0.6023 &  0.7755 & 0.6488 & 0.7341 \\
 $Q_1$, $P_2$ & 0.6302 & 0.5988 & 0.5343 &  0.7764 & 0.7207 & 0.6854 \\
 $Q_1$, $P_1$ & 0.7549 & 0.4968 & 0.5078 &  0.8779 & 0.6385 & 0.6629 \\
 $P_3$, $P_2$ & 0.6108 & 0.8456 & 0.8583 &  0.7330 & 0.7637 & 0.7817 \\
 $P_3$, $P_1$ & 0.8102 & 0.4944 & 0.7059 &  0.8471 & 0.6036 & 0.7590 \\
 $P_2$, $P_1$ & 0.8065 & 0.6963 & 0.4919 &  0.8454 & 0.7481 & 0.5741 \\
 $Q_3$, $Q_2$, $Q_1$ & 0.5902 & 0.4941 & 0.5501 & 0.7899 & 0.6566 & 0.7202 \\
 $Q_3$, $Q_2$, $P_3$ & 0.6072 & 0.5351 & 0.6046 & 0.7635 & 0.6667 & 0.7376 \\
 $Q_3$, $Q_2$, $P_2$ & 0.6101 & 0.5912 & 0.5629 & 0.7614 & 0.7339 & 0.6940 \\
 $Q_3$, $Q_2$, $P_1$ & 0.6986 & 0.5144 & 0.5577 & 0.8508 & 0.6394 & 0.6857 \\
 $Q_3$, $Q_1$, $P_3$ & 0.6177 & 0.5405 & 0.5851 & 0.7826 & 0.6695 & 0.7583 \\
 $Q_3$, $Q_1$, $P_2$ & 0.6400 & 0.6358 & 0.5318 & 0.7919 & 0.7631 & 0.7136 \\
 $Q_3$, $Q_1$, $P_1$ & 0.7532 & 0.4911 & 0.5272 & 0.8790 & 0.6522 & 0.7067 \\
 $Q_3$, $P_3$, $P_2$ & 0.6239 & 0.8500 & 0.7952 & 0.7356 & 0.8219 & 0.7660 \\
 $Q_3$, $P_3$, $P_1$ & 0.8069 & 0.4988 & 0.6968 & 0.8548 & 0.6122 & 0.7529 \\
 $Q_3$, $P_2$, $P_1$ & 0.8023 & 0.6937 & 0.5076 & 0.8503 & 0.7523 & 0.6437 \\
 $Q_2$, $Q_1$, $P_3$ & 0.6329 & 0.5124 & 0.6436 & 0.8043 & 0.6743 & 0.7861 \\
 $Q_2$, $Q_1$, $P_2$ & 0.6292 & 0.5812 & 0.5465 & 0.7978 & 0.7328 & 0.6938 \\
 $Q_2$, $Q_1$, $P_1$ & 0.7612 & 0.5049 & 0.5232 & 0.8811 & 0.6683 & 0.6797 \\
 $Q_2$, $P_3$, $P_2$ & 0.6059 & 0.7882 & 0.8550 & 0.7607 & 0.7520 & 0.8307 \\
 $Q_2$, $P_3$, $P_1$ & 0.8055 & 0.5029 & 0.7119 & 0.8524 & 0.6340 & 0.7709 \\
 $Q_2$, $P_2$, $P_1$ & 0.8169 & 0.6692 & 0.5051 & 0.8576 & 0.7400 & 0.6165 \\
 $Q_1$, $P_3$, $P_2$ & 0.6354 & 0.8278 & 0.8265 & 0.7876 & 0.7660 & 0.7731 \\
 $Q_1$, $P_3$, $P_1$ & 0.8070 & 0.5115 & 0.6879 & 0.8700 & 0.6476 & 0.7670 \\
 $Q_1$, $P_2$, $P_1$ & 0.8049 & 0.6871 & 0.5168 & 0.8662 & 0.7595 & 0.6538 \\
 $P_3$, $P_2$, $P_1$ & 0.8067 & 0.8150 & 0.8229 & 0.8487 & 0.7658 & 0.7633 \\
 $Q_3$, $Q_2$, $Q_1$, $P_3$ & 0.6289 & 0.5288 & 0.6266 & 0.8076 & 0.6904 & 0.8054 \\
 $Q_3$, $Q_2$, $Q_1$, $P_2$ & 0.6365 & 0.6110 & 0.5596 & 0.8077 & 0.7937 & 0.7322 \\
 $Q_3$, $Q_2$, $Q_1$, $P_1$ & 0.7537 & 0.5016 & 0.5509 & 0.8893 & 0.6819 & 0.7169 \\
 $Q_3$, $Q_2$, $P_3$, $P_2$ & 0.6162 & 0.8627 & 0.8549 & 0.7631 & 0.8439 & 0.8303 \\
 $Q_3$, $Q_2$, $P_3$, $P_1$ & 0.8086 & 0.5132 & 0.6983 & 0.8581 & 0.6552 & 0.7652 \\
 $Q_3$, $Q_2$, $P_2$, $P_1$ & 0.8138 & 0.6736 & 0.5337 & 0.8574 & 0.7631 & 0.6761 \\
 $Q_3$, $Q_1$, $P_3$, $P_2$ & 0.6424 & 0.8407 & 0.7974 & 0.7949 & 0.8245 & 0.8028 \\
 $Q_3$, $Q_1$, $P_3$, $P_1$ & 0.8091 & 0.5181 & 0.6845 & 0.8729 & 0.6630 & 0.7847 \\
 $Q_3$, $Q_1$, $P_2$, $P_1$ & 0.8083 & 0.6830 & 0.5379 & 0.8686 & 0.7800 & 0.7047 \\
 $Q_3$, $P_3$, $P_2$, $P_1$ & 0.8054 & 0.8194 & 0.7916 & 0.8514 & 0.8050 & 0.7969 \\
 $Q_2$, $Q_1$, $P_3$, $P_2$ & 0.6359 & 0.8010 & 0.8413 & 0.8105 & 0.7784 & 0.8351 \\
 $Q_2$, $Q_1$, $P_3$, $P_1$ & 0.8104 & 0.5246 & 0.7019 & 0.8735 & 0.6765 & 0.7940 \\
 $Q_2$, $Q_1$, $P_2$, $P_1$ & 0.8219 & 0.6682 & 0.5249 & 0.8762 & 0.7607 & 0.6734 \\
 $Q_2$, $P_3$, $P_2$, $P_1$ & 0.8133 & 0.7799 & 0.8236 & 0.8561 & 0.7863 & 0.8126 \\
 $Q_1$, $P_3$, $P_2$, $P_1$ & 0.8093 & 0.8070 & 0.8024 & 0.8633 & 0.7673 & 0.7711 \\
 $Q_3$, $Q_2$, $Q_1$, $P_3$, $P_2$ & 0.6387 & 0.8516 & 0.8452 & 0.8109 & 0.8333 & 0.8406 \\
 $Q_3$, $Q_2$, $Q_1$, $P_3$, $P_1$ & 0.8164 & 0.5283 & 0.6906 & 0.8796 & 0.6936 & 0.7987 \\
 $Q_3$, $Q_2$, $Q_1$, $P_2$, $P_1$ & 0.8213 & 0.6673 & 0.5517 & 0.8775 & 0.7907 & 0.7172 \\
 $Q_3$, $Q_2$, $P_3$, $P_2$, $P_1$ & 0.8133 & 0.8496 & 0.8457 & 0.8571 & 0.8300 & 0.8187 \\
 $Q_3$, $Q_1$, $P_3$, $P_2$, $P_1$ & 0.8103 & 0.8190 & 0.7974 & 0.8658 & 0.8197 & 0.8138 \\
 $Q_2$, $Q_1$, $P_3$, $P_2$, $P_1$ & 0.8218 & 0.7905 & 0.8176 & 0.8696 & 0.7923 & 0.8236 \\
 $Q_3$, $Q_2$, $Q_1$, $P_3$, $P_2$, $P_1$ & 0.8197 & 0.8451 & 0.8408 & 0.8734 & 0.8255 & 0.8276 \\
\hline 
  \caption{The sensitivities achieved across the IVBC combination searches by the LR and SVM classification methods.}
 \label{table_sens_IVBC}
\end{longtable}
\clearpage

\begin{longtable}{|c|c c c | c c c|}

\hline
& \multicolumn{3}{c|}{\textbf{LR}} & \multicolumn{3}{c|}{\textbf{SVM}}\\
\textbf{Input combination} & \textbf{Aortic} & \textbf{Iliac 1} & \textbf{Iliac 2} & \textbf{Aortic} & \textbf{Iliac 1} & \textbf{Iliac 2} \\
\hline
\endhead
\hline
\endfoot
\endlastfoot
 $Q_3$ & 0.4688 & 0.6165 & 0.5352 & 0.5250 & 0.7410 & 0.6320 \\
 $Q_2$ & 0.4382 & 0.5448 & 0.4703 & 0.4915 & 0.5879 & 0.6371 \\
 $Q_1$ & 0.4703 & 0.4602 & 0.4777 & 0.6405 & 0.7463 & 0.7023 \\
 $P_3$ & 0.4652 & 0.4989 & 0.5133 & 0.5633 & 0.6781 & 0.6195 \\
 $P_2$ & 0.4580 & 0.5150 & 0.4793 & 0.5735 & 0.6976 & 0.6260 \\
 $P_1$ & 0.5255 & 0.5070 & 0.4871 & 0.6300 & 0.6899 & 0.6799 \\
 $Q_3$, $Q_2$ & 0.4605 & 0.5502 & 0.4956 &  0.5619 & 0.7514 & 0.7144 \\
 $Q_3$, $Q_1$ & 0.5026 & 0.5023 & 0.5166 &  0.6934 & 0.8042 & 0.7401 \\
 $Q_3$, $P_3$ & 0.4638 & 0.5139 & 0.5242 &  0.6341 & 0.7365 & 0.6828 \\
 $Q_3$, $P_2$ & 0.4658 & 0.5150 & 0.5312 &  0.6366 & 0.7153 & 0.7202 \\
 $Q_3$, $P_1$ & 0.5184 & 0.5181 & 0.4842 &  0.6857 & 0.7448 & 0.6985 \\
 $Q_2$, $Q_1$ & 0.4793 & 0.5225 & 0.4734 &  0.6729 & 0.7937 & 0.7428 \\
 $Q_2$, $P_3$ & 0.4500 & 0.5623 & 0.5041 &  0.5949 & 0.7574 & 0.6711 \\
 $Q_2$, $P_2$ & 0.4581 & 0.5348 & 0.4718 &  0.5893 & 0.7056 & 0.6780 \\
 $Q_2$, $P_1$ & 0.5189 & 0.5281 & 0.4703 &  0.6652 & 0.7109 & 0.7049 \\
 $Q_1$, $P_3$ & 0.4791 & 0.4947 & 0.5377 &  0.7312 & 0.8260 & 0.7651 \\
 $Q_1$, $P_2$ & 0.4917 & 0.5571 & 0.5258 &  0.7307 & 0.7856 & 0.7544 \\
 $Q_1$, $P_1$ & 0.5487 & 0.5256 & 0.5265 &  0.7628 & 0.8132 & 0.7849 \\
 $P_3$, $P_2$ & 0.4521 & 0.6623 & 0.6576 &  0.5865 & 0.7214 & 0.6797 \\
 $P_3$, $P_1$ & 0.5700 & 0.6718 & 0.5194 &  0.6899 & 0.7400 & 0.6358 \\
 $P_2$, $P_1$ & 0.5771 & 0.5179 & 0.6370 &  0.6927 & 0.6573 & 0.7430 \\
 $Q_3$, $Q_2$, $Q_1$ & 0.4931 & 0.5235 & 0.5032 & 0.7038 & 0.8308 & 0.7795 \\
 $Q_3$, $Q_2$, $P_3$ & 0.4582 & 0.5420 & 0.5137 & 0.6489 & 0.7805 & 0.7247 \\
 $Q_3$, $Q_2$, $P_2$ & 0.4601 & 0.5363 & 0.5153 & 0.6496 & 0.7383 & 0.7596 \\
 $Q_3$, $Q_2$, $P_1$ & 0.5214 & 0.5146 & 0.4805 & 0.7065 & 0.7711 & 0.7323 \\
 $Q_3$, $Q_1$, $P_3$ & 0.4902 & 0.4900 & 0.5656 & 0.7481 & 0.8271 & 0.7855 \\
 $Q_3$, $Q_1$, $P_2$ & 0.4885 & 0.5397 & 0.5501 & 0.7456 & 0.7972 & 0.8045 \\
 $Q_3$, $Q_1$, $P_1$ & 0.5524 & 0.5380 & 0.5481 & 0.7727 & 0.8278 & 0.7991 \\
 $Q_3$, $P_3$, $P_2$ & 0.4522 & 0.6766 & 0.7246 & 0.6412 & 0.7213 & 0.7837 \\
 $Q_3$, $P_3$, $P_1$ & 0.5963 & 0.6736 & 0.5452 & 0.7127 & 0.7767 & 0.6986 \\
 $Q_3$, $P_2$, $P_1$ & 0.5742 & 0.5445 & 0.6277 & 0.7087 & 0.7104 & 0.7759 \\
 $Q_2$, $Q_1$, $P_3$ & 0.4807 & 0.5569 & 0.5180 & 0.7275 & 0.8349 & 0.7768 \\
 $Q_2$, $Q_1$, $P_2$ & 0.4833 & 0.5823 & 0.5051 & 0.7259 & 0.7907 & 0.7772 \\
 $Q_2$, $Q_1$, $P_1$ & 0.5553 & 0.5517 & 0.4920 & 0.7654 & 0.8239 & 0.7870 \\
 $Q_2$, $P_3$, $P_2$ & 0.4499 & 0.7278 & 0.6786 & 0.6109 & 0.7937 & 0.6791 \\
 $Q_2$, $P_3$, $P_1$ & 0.5744 & 0.6717 & 0.5240 & 0.7090 & 0.7882 & 0.6779 \\
 $Q_2$, $P_2$, $P_1$ & 0.6012 & 0.5550 & 0.5956 & 0.7030 & 0.7050 & 0.7576 \\
 $Q_1$, $P_3$, $P_2$ & 0.4910 & 0.6625 & 0.6649 & 0.7236 & 0.7956 & 0.7697 \\
 $Q_1$, $P_3$, $P_1$ & 0.5935 & 0.6307 & 0.5409 & 0.7633 & 0.8193 & 0.7503 \\
 $Q_1$, $P_2$, $P_1$ & 0.6086 & 0.5670 & 0.6002 & 0.7662 & 0.7647 & 0.7998 \\
 $P_3$, $P_2$, $P_1$ & 0.5862 & 0.6977 & 0.6933 & 0.6918 & 0.7235 & 0.7045 \\
 $Q_3$, $Q_2$, $Q_1$, $P_3$ & 0.4847 & 0.5449 & 0.5435 & 0.7519 & 0.8467 & 0.8018 \\
 $Q_3$, $Q_2$, $Q_1$, $P_2$ & 0.4962 & 0.5774 & 0.5252 & 0.7521 & 0.7891 & 0.8157 \\
 $Q_3$, $Q_2$, $Q_1$, $P_1$ & 0.5685 & 0.5562 & 0.5154 & 0.7814 & 0.8360 & 0.8150 \\
 $Q_3$, $Q_2$, $P_3$, $P_2$ & 0.4518 & 0.7727 & 0.7614 & 0.6608 & 0.7996 & 0.7961 \\
 $Q_3$, $Q_2$, $P_3$, $P_1$ & 0.5898 & 0.6600 & 0.5434 & 0.7270 & 0.8114 & 0.7383 \\
 $Q_3$, $Q_2$, $P_2$, $P_1$ & 0.6020 & 0.5693 & 0.5993 & 0.7271 & 0.7436 & 0.7916 \\
 $Q_3$, $Q_1$, $P_3$, $P_2$ & 0.4922 & 0.6805 & 0.7225 & 0.7488 & 0.8107 & 0.8244 \\
 $Q_3$, $Q_1$, $P_3$, $P_1$ & 0.6100 & 0.6309 & 0.5653 & 0.7752 & 0.8381 & 0.7874 \\
 $Q_3$, $Q_1$, $P_2$, $P_1$ & 0.6096 & 0.5688 & 0.6038 & 0.7780 & 0.7930 & 0.8169 \\
 $Q_3$, $P_3$, $P_2$, $P_1$ & 0.6017 & 0.7082 & 0.7338 & 0.7141 & 0.7619 & 0.7821 \\
 $Q_2$, $Q_1$, $P_3$, $P_2$ & 0.4921 & 0.7347 & 0.6891 & 0.7296 & 0.8405 & 0.7843 \\
 $Q_2$, $Q_1$, $P_3$, $P_1$ & 0.5978 & 0.6345 & 0.5418 & 0.7713 & 0.8381 & 0.7730 \\
 $Q_2$, $Q_1$, $P_2$, $P_1$ & 0.6119 & 0.5873 & 0.5798 & 0.7707 & 0.7893 & 0.8022 \\
 $Q_2$, $P_3$, $P_2$, $P_1$ & 0.6012 & 0.7425 & 0.7040 & 0.7074 & 0.7954 & 0.7248 \\
 $Q_1$, $P_3$, $P_2$, $P_1$ & 0.6096 & 0.6850 & 0.6861 & 0.7612 & 0.7895 & 0.7702 \\
 $Q_3$, $Q_2$, $Q_1$, $P_3$, $P_2$ & 0.4856 & 0.7612 & 0.7505 & 0.7516 & 0.8464 & 0.8409 \\
 $Q_3$, $Q_2$, $Q_1$, $P_3$, $P_1$ & 0.6137 & 0.6308 & 0.5653 & 0.7821 & 0.8563 & 0.8090 \\
 $Q_3$, $Q_2$, $Q_1$, $P_2$, $P_1$ & 0.6212 & 0.5825 & 0.5870 & 0.7862 & 0.8045 & 0.8354 \\
 $Q_3$, $Q_2$, $P_3$, $P_2$, $P_1$ & 0.6010 & 0.7878 & 0.7677 & 0.7281 & 0.8103 & 0.8172 \\
 $Q_3$, $Q_1$, $P_3$, $P_2$, $P_1$ & 0.6177 & 0.7093 & 0.7338 & 0.7755 & 0.8226 & 0.8266 \\
 $Q_2$, $Q_1$, $P_3$, $P_2$, $P_1$ & 0.6128 & 0.7475 & 0.7032 & 0.7665 & 0.8352 & 0.7964 \\
 $Q_3$, $Q_2$, $Q_1$, $P_3$, $P_2$, $P_1$ & 0.6225 & 0.7859 & 0.7614 & 0.7847 & 0.8479 & 0.8488 \\
\hline 
  \caption{The specificities achieved across the IVBC combination searches by the LR and SVM classification methods.}
 \label{table_spec_IVBC}
\end{longtable}

\end{document}